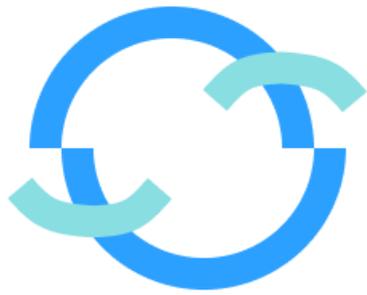

# Fidelity and Privacy of Synthetic Medical Data

## Review of Methods and Experimental Results

June 2021

Ofer Mendelevitch, Michael D. Lesh SM MD FACC,

# Table of Contents

## Abstract

The digitization of medical records ushered in a new era of big data to clinical science, and with it the possibility that data could be shared, to multiply insights beyond what investigators could abstract from paper records. The need to share individual-level medical data to accelerate innovation in precision medicine continues to grow, and has never been more urgent, as scientists grapple with the COVID-19 pandemic. However, enthusiasm for the use of big data has been tempered by a fully appropriate concern for patient autonomy and privacy. That is, the ability to extract private or confidential information about an individual, in practice, renders it difficult to share data, since significant infrastructure and data governance must be established before data can be shared. Although HIPAA provided de-identification as an approved mechanism for data sharing, linkage attacks were identified as a major vulnerability. A variety of mechanisms have been established to avoid leaking private information, such as field suppression or abstraction, strictly limiting the amount of information that can be shared, or employing mathematical techniques such as differential privacy. Another approach, which we focus on here, is creating synthetic data that mimics the underlying data. For synthetic data to be a useful mechanism in support of medical innovation and a proxy for real-world evidence, one must demonstrate two properties of the synthetic dataset: (1) any analysis on the real data must be matched by analysis of the synthetic data (statistical fidelity) and (2) the synthetic data must preserve privacy, with minimal risk of re-identification (privacy guarantee). In this paper we propose a framework for quantifying the statistical fidelity and privacy preservation properties of synthetic datasets and demonstrate these metrics for synthetic data generated by Syntegra technology.

## 1. Introduction

A prerequisite to healthcare innovation is the availability of high-quality, unbiased, and diverse patient-level medical datasets. Increasingly, patient data from patient care and clinical trials or an increasing number of commercial data sources (for example, there is now a wealth of general consumer data as well as specific health-related data from wearable devices and fitness apps) are being generated by providers, governments, industry and individuals themselves. While such datasets can be a rich resource for investigators in those organizations, they are generally not accessible to the broader research community due to patient privacy concerns. Even when it is possible for a researcher to gain access to such data, ensuring proper governance and complying with strict legal requirements is a lengthy and expensive process. This can severely hamper the timeliness of research and, consequently, its translational benefits to patient care. This delay is particularly devastating now, during the rapidly advancing COVID-19 pandemic.

As noted in [1], "sharing data produced from clinical trials...has the potential to advance scientific discovery, improve clinical care, and increase knowledge gained from data collected in these trials. As such, data sharing has become an ethical and scientific imperative." Unfortunately, this ethical mandate may conflict with the equally important ethical and legal mandate to protect patient privacy, and the reality is that data owners are reluctant to share patient level medical datasets. And even with these restrictions, "data sharing" as described in the academic literature generally applies to credentialed academic researchers collaborating with other academic investigators, or life-sciences industries who can provide large research grants to those academics.

As discussed in Section 1.1 below, methods of de-identification that are based on HIPAA's safe-harbor provisions ([2]), first established by law in 1996, are much less effective now, considering the vulnerability to linkage attacks - combining "de-identified" data with information available elsewhere, such as social media or public records, to extract highly sensitive personal information ([3], [4]). At the same time, they may also significantly degrade the utility of the data [25].

### 1.1 De-Identification and Re-Identification

As detailed in [5], the simplest method of privacy protection is to remove all fields (HIPAA prescribed 18 such fields) that could directly and uniquely identify an individual, such as name, social security number, and phone number. Until the mid 2000s, this was considered adequate to de-identify data, and pursuant to the HIPAA Privacy Rule 45 CFR 164 such data no longer constitutes protected health information (PHI). The assumption was that once these 18 types of information were removed or masked, and the disclosing entity had no actual knowledge that the information in the de-identified dataset could be used to identify an individual, the disclosure did not constitute a significant risk to privacy. However, in recent years, it became clear that many other data fields can be used to identify individuals [4].

GDPR expands the scope of protected information beyond PHI, instead using the term "personal data," where "personal data" means *any* information relating to an identified or identifiable natural person ("data subject"), whereas HIPAA is limited to information generated by *healthcare providers*, insurers and clearinghouses and pertaining to the medical treatment of patients. An identifiable natural person is one who can be identified, directly or indirectly, in particular by reference to an identifier such as a name, an identification number, location data, an online identifier or to one or more factors specific to the physical, physiological, genetic, mental, economic, cultural or social identity of that natural person.

Moreover, GDPR distinguishes two types of de-identification: pseudonymization and anonymization. Pseudonymized data may in many cases be similar to de-identified data under HIPAA, but contrary to HIPAA which explicitly does not apply to de-identified data, GDPR still imposes legal restrictions on pseudonymized data, albeit less stringent ones than for fully identifiable data. For example, data containing an encrypted patient key could potentially still be rendered de-identified under HIPAA, provided only the data source, but not the data recipient can decrypt this key or otherwise use it to recover the patient's identity, whereas under GDPR such data would only be pseudonymized, and GDPR would continue to apply. Full anonymization pursuant to GDPR is defined such that even if the identified source data or any other auxiliary data (whether or not available to the data recipient), could be used to recover patient identity; it is, however, difficult to achieve and retains very little analytic utility. Thus, in order to be able to use GDPR protected data, significant legal burdens will typically have to be met, and this can be especially challenging for entities outside the EU. Importantly, defining "*identifiable*" data is complicated by the possibility of re-identification [6]. Re-identification is the matching of anonymized data back to an individual. In recent years, faith that de-identification prevents re-identification can no longer be supported. For example, a 2009 Social Security Number study [7] showed that data about an individual's place and date of birth, voter registration, and other publicly available information can be used to predict their Social Security number.

Traditional techniques for disclosure limitation can be classified in a number of ways, grouped by information limiting methods and data perturbation methods. Information limiting methods are those that delete, mask, suppress, or obscure data fields or values in order to prevent re-identification. Data perturbation methods are those that use statistical means to alter the underlying data itself, by adding noise and/or limit the query results that can be drawn from the data.

## 1.2 Synthetic Data

In this paper, we will focus on the use of synthetic medical data, a novel method for data sharing that does not explicitly limit or perturb data, and instead learns a high granularity statistical representation from the data in order to generate completely synthetic medical records using sampling and randomization. This in principle can provide high statistical fidelity and low risk of disclosure.

Given the stakes in healthcare, when claims are made that any particular method renders a dataset non-re-identifiable, there must be a set of metrics to ensure that such a claim is warranted. In general, there is a tradeoff between the risk of disclosure and the utility of a disclosed dataset - the higher the utility, the lower the level of privacy guarantee, and vice-versa, as shown in figure 1:

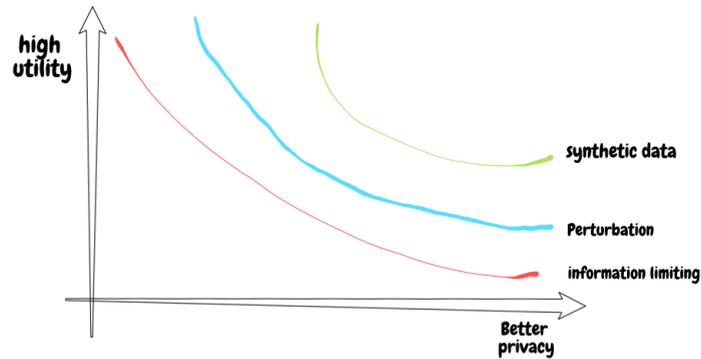

**Figure 1:** trade-off between utility and privacy

Different disclosure techniques represent different trade-off curves; for example, information-limiting techniques are often more detrimental to utility than perturbation techniques, whereas synthetic data (if done properly) can provide the best trade-off.

Therefore, it is essential to develop a set of metrics and associated acceptance criteria that can be used to determine when synthetic data can be trusted to provide real world evidence. Of course, any real-world dataset from which an investigator hopes to draw conclusions based on a statistical analysis, is itself a sample of a much larger universe. In other words, our expectation that the synthetic dataset matches the real dataset is really a proxy for the expectation that the synthetic dataset and real dataset are both samples from the same underlying distribution. As an example, suppose we have a dataset of 1000 patients with coronary heart disease, 500 of whom have had treatment X and 500 have been treated with a placebo. The outcome is that treatment X is significantly better than placebo. A clinician or a regulatory body has enough assurance that they are willing to recommend treatment X for all qualifying patients with heart disease, acknowledging that those 1000 patients are a representative sample of say 19,000,000 Americans with coronary heart disease. Taken together, the fidelity metrics suggested in this paper provide a high level of assurance that the real dataset and synthetic dataset are drawn from the same distribution.

In fact, we may find that synthetic data is actually a better representation of the ground truth. That is because creation of the synthetic dataset may in some cases remove out of distribution anomalies. The synthetic data engine could have an error filtering function that is not present when the real dataset is used directly in statistical analyses.

We present here a collection of metrics and visualizations to assess the privacy and fidelity claims of any method for synthetic data generation, and the method of synthetic medical data generation in particular. In section 2 we provide a brief overview of synthetic data generation in healthcare. In section 3 and 4 we describe our proposed framework for evaluating statistical fidelity of synthetic data against the real data and measuring privacy and disclosure risk. In section 5 we present experimental results with four datasets, using Syntegra's generative engine, in section 6 we analyze our results, and in section 7 we present our conclusions.

## 2. Synthetic Data in Medicine

Synthetic data has been used in a number of applications, such as computer vision and robotics [8], and creating synthetic controls for economics and social sciences [9], where randomized trials are not possible, but a comparison to an intervention is required. The privacy preserving properties of synthetic data has been initially discussed in [10], and much later in [11] and [21].

For reducing the risk of patient data re-identification and accelerating the process by which such data is made more widely available, synthetically generated data is a promising alternative or addition to standard anonymization procedures. Some mechanisms of creating synthetic data do so based on simulating disease processes, known patterns of care, or disease-specific guidelines (notably [12] and [13]), which means starting with heuristics about medical conditions, such as practice guidelines or literature review, and letting rules drive new record production. These methods typically fail to anticipate unusual edge cases, such as surgical complications, other adverse events or rare conditions, which means that they fail to satisfy the needs of modern precision medicine. For regulatory approval or post-approval vigilance of therapeutics that require real world data, and where synthetic data will substitute for the actual data because of privacy concerns, the synthetic data should be created from the data itself [14], i.e., data-driven synthesis as opposed to process- or rule-based simulation. Indeed, outside of pedagogical or population-based modeling, synthetic data has seen limited use in mission-critical applications such as synthetic control groups in clinical trials to support regulatory approval or post-approval surveillance.

## 2.1 The Syntegra Synthetic Data Engine and Medical Mind

At Syntegra, we have developed a novel machine-learning-based synthetic data generator, the Syntegra Medical Mind, that can convert any type of clinical data into an equivalent synthetic version. The patient-level data can be from any real-world database (RWD), or from observational or prospective clinical trials. The synthetic dataset is intended to match the statistical properties of the original, while providing strong privacy protection, as validated by the metrics defined below. Note that rather than providing only aggregate statistics, the synthetic data is created at the level of individual participants. The goal is a dataset that mimics the statistical properties of the real data and can be used for any analysis, including training-state-of-the-art, predictive models like random forest or gradient boosted trees, with performance that matches that of models trained on the original data.

Unlike many previous synthetic data techniques that add noise or attempt to simulate results based on hand-coded rules or guidelines, our method is data-centric: it uses a very large neural network to learn, in an unsupervised fashion, the underlying probability distribution in the real data. A synthetic dataset is then generated that accurately maintains the statistical properties of the real data, while preserving privacy. The engine works by viewing all the data points for a given participant as a "patient sentence", with events in time or tabular data, and learning the underlying latent probability distribution by training a language models[1] on those "sentences"; subsequently, the trained model can be used to generate synthetic sentences by sampling from the learned distribution (which is encoded by the neural network). By repeating that process any number of times, we generate a set of patient-level clinical records. Utilizing transfer learning, our generative model leverages a pre-trained corpus that is fine-tuned, in an unsupervised fashion, by the real dataset that one wants converted into a synthetic equivalent. The pre-trained corpus contains a representation of medical patterns extracted from general and health-care related data, augmented by new datasets as they are encountered. That is, the Syntegra Medical Mind improves over time as each new set of real data enriches the pre-trained corpus.

# 3. Statistical Fidelity Validation

Given a real medical dataset R, and a synthetic dataset S generated based on R, our goal is to measure the statistical fidelity between R and S. But what do we mean by the term "statistical fidelity"?

Here, we explore six methods to compare the degree to which the synthetic dataset is an accurate replica of the original dataset:

1) Visualize the real and synthetic datasets using dimensionality reduction
2) Compare summary statistics
3) Compare single variable distributions
4) Compare pairwise correlations
5) Multivariate and non-linear metrics
6) Clinical consistency check on the synthetic data

As demonstrated by the famous Anscombe's quartet [15] example - statistical metrics have known limitations, which proper visualization can sometimes help address. Thus, in our framework for statistical fidelity we use a combination of visualization and metrics, recognizing that good metrics provide an easy numerical value by which to judge fidelity comparing real to synthetic data, whereas visualization provides a more nuanced view of differences.

## 3.1 Record Distance Metric

Some statistical fidelity and privacy tests require a reasonable measure of distance between each pair of records. Some common distance functions are Euclidean distance, cosine distance, Gower distance, or Hamming distance; many others exist - see e.g. [17]). With healthcare data it is important to provide a true distance metric[2] that appropriately deals with both numeric and categorical values, as well as missing values.

---

[1] https://openai.com/blog/better-language-models/
[2] A true distance metric is symmetric and satisfies the triangle inequality, with each item having distance 0 to itself.

Here we describe two possible choices for a distance metric. The first one is Gower distance, which was first described in [30] and is common for data with mixed (both numeric and categorical) variables, and a robust extension of Gower distance, replacing normalized manhattan distance with a variant of wave-hedges is described in [29].

Given two vectors $x_i$ and $x_j$, both with dimension p, the enhanced gower distance is defined as:

$$\text{Eq (1): } d_G(x_i, x_j) = \frac{\sum_{m=1}^{p} w_{ijm} d_{ijm}}{\sum_{m=1}^{p} w_{ijm}}$$

where $d_{ijm}$ is the distance between the m-th variable of x and y, and $w_{ijm}$ is an optional weighting for each variable. In this paper we use $w_{ijm} = 1$ for all variables, but we note that an interesting alternative to consider is weighting variables by their relative importance in the use of the distance metric.

Based on combining [29] and [30] we define the individual distance for a given variable as follows:

- For categorical variables: $d_{ijm} = \begin{cases} 0, \text{ for } x_i = x_j \\ 1, \text{ for } x_i \neq x_j \end{cases}$

- For numerical variables, $d_{ijm} = \begin{cases} 1 - \frac{1 + min(x_i, x_j)}{1 + max(x_i, x_j)}, & \text{for } min(x_i, x_j) \geq 0 \\ 1 - \frac{1}{1 + max(x_i, x_j) + |min(x_i, x_j)|}, & \text{for } min(x_i, x_j) < 0 \end{cases}$

An alternative approach, which we've used to report the results in sections 4 and 5 below, is to bucket numeric values into percentile bins, and in that way transform them into categorical variables. The distance computation in this case remains the same, and we only have categorical variables.

Lastly, to deal properly with missing values, we use the following approach:
- For categorical values, we define a "missing value" as a separate (additional) category
- For numerical values, we define $d_{ijm} = 1$ if one of $x_{im}$ or $x_{jm}$ is missing but the other is not, and as 0 if both are missing.

## 3.2 Visualize and Compare Datasets
Before diving into numerical analysis, it is an accepted data science methodology first to visualize datasets, especially where one wishes to compare two datasets, and where the number of records is large. Graphical visualization makes use of human's innate ability to recognize patterns, concordance, and deviation. However, given that a medical dataset can be of very high dimensionality, with heterogeneous data types, it can be a challenge to create a visualization that is coherent to a human observer. Recently, robust dimensionality reduction methods such as tSNE and UMAP [22] were introduced. UMAP seeks to learn the manifold structure of a dataset and find a low dimensional embedding that preserves the essential topological structure of that manifold.

In Figure 2a, each point represents a dimensionally reduced single patient record from a high-dimensional medical dataset. The axes themselves have no convenient interpretation, but the clustering present in the data is clear in the visualization. Figure 2a shows the full UMAP scatter plot of the whole datasets, whereas figure 2b zooms in on a specific small cluster of points in more detail.

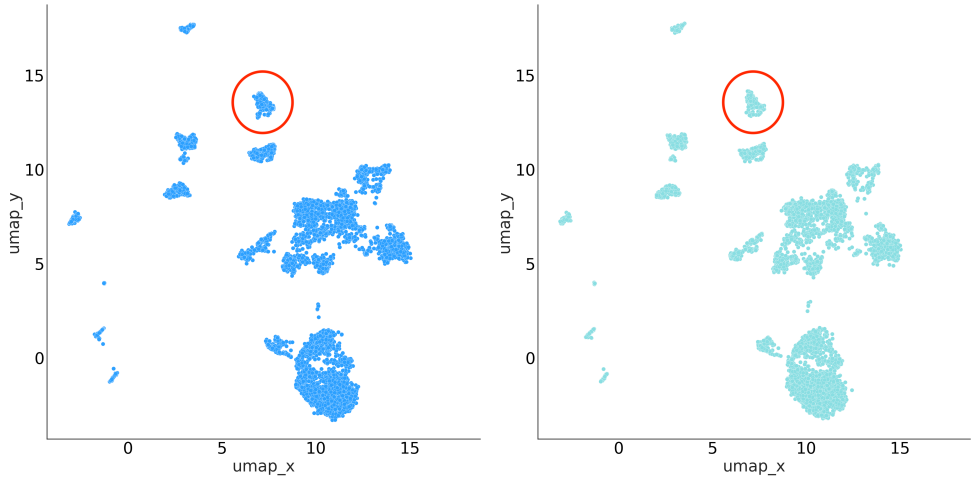

**Figure 2a**: UMAP visualization of a clinical trial dataset. On left, the real dataset; on right, the synthetic dataset.

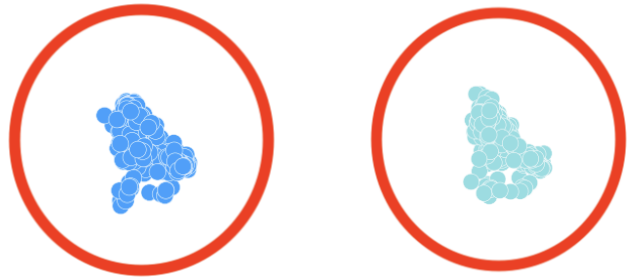

**Figure 2b**: UMAP visualization of a clinical trial dataset - small cluster of patient records (zoomed in) - real data on the left, synthetic data on the right

Undoubtedly, some records are close to others, hence clustering on multiple scales is noted. Some records are "edge cases" and represent small groups far from the centroid (red circles). Note that the coverage of the synthetic dataset (figure 2a - right) is quite an accurate representation of the distribution of the real data (figure 2a - left), including covering the edge cases. This is important, as simple numerical analysis may show that the synthetic and real data have similar central tendency, which might miss a failure of the synthetic dataset to replicate the small cohorts or edge cases present in the real data. The distribution of real and synthetic points is ideally close but should not be identical. That is, the synthetic data should not be just a copy of the real. Larger magnification reveals that they are not identical, and our privacy metrics (see section 4) prove that copying has not taken place.

By using UMAP, we can clearly see that even small cohorts are picked up in the synthetic data. Another reason small cohort coverage is important is that standard de-identification methods, such as for HIPAA compliant de-identification, often require removal of small cohorts out of concern for re-identification via membership inference. So, a truly robust synthetic data engine will maintain these small cohorts, while still maintaining privacy.

### 3.3 Population Statistics

The next step in validation that should be performed when presented with any new dataset is summary statistics or population statistics [16]. The difference in our case is that these summary statistics will be computed in a paired fashion on the real and then on the synthetic data, and then compared. Table 1a provides an example for numeric variables, where we compare the mean and standard deviation for real vs synthetic, whereas table 1b provides an example for categorical variables, where we compare the count and percentage in each category:

**Numeric variables**

| variable | Real | | Synthetic | |
|---|---|---|---|---|
| | mean | SD | mean | SD |

**Categorical variables**

| variable | Value | Real | Synthetic |
|---|---|---|---|

| | | | | |
|---|---|---|---|---|
| AGE | 55.4 | 8.8 | 55.3 | 8.9 |
| BMI | 24.3 | 4.9 | 24.5 | 5.0 |
| HEIGHT | 170.1 | 22.1 | 171.2 | 21.9 |
| WEIGHT | 82.9 | 23.5 | 83.2 | 23.5 |

Table 1a

| | | n | % | n | % |
|---|---|---|---|---|---|
| SEX | Female | 1,519 | 22.3 | 1634 | 24.0 |
| | Male | 5,281 | 77.7 | 5166 | 76.0 |
| RACE | NonWhite | 991 | 14.6 | 1073 | 15.8 |
| | White | 5809 | 85.4 | 5727 | 84.2 |

Table 1b

The goal of this exercise is to understand whether population-level statistics on the real data match those of the synthetic data.

### 3.4 Single Variable (Marginal) Distributions

Moving from variable-level statistics to distributions we compare the *distribution of a given variable* in the real data to its distribution in the synthetic data. For numeric variables, the distributions can be easily compared using a histogram:

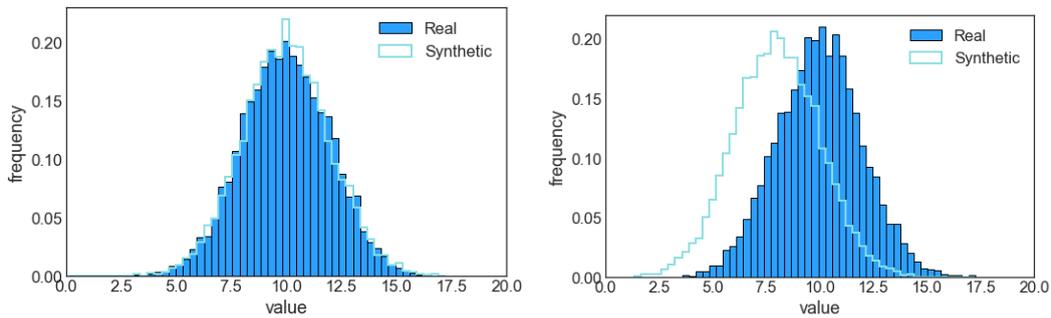

**Figures 3a and 3b:** Histogram comparing similar (left) and differing (right) distributions of numeric variables

The (non-parametric) 2-sided KS-statistic[3] can be used as a statistical test to determine whether the two variables (real vs. synthetic) are drawn from the same distribution. The KS-statistic is a value between 0 and 1; when this value is small (or the associated p-value is high, above 0.05), then we cannot reject the null hypothesis that the distribution of the real variable is the same as that of the synthetic variable, which means there is high statistical fidelity of synthetic data when compared to real data.

For the example in figure 3a above, the KS-statistic value is 0.0126 and the p-value is 0.405, clearly consistent with a good fit. Figure 3b demonstrates the opposite case where the two distributions don't match -In this case the KS-statistic value is 0.392 and the p-value<0.0001, again consistent with our expectations of a poor fit.

For categorical variables, the *category-based histogram* is a useful visualization tool to understand differences in distribution:

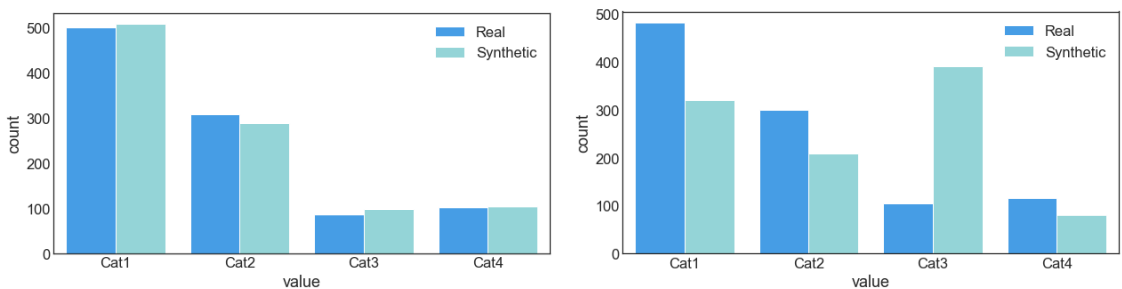

**Figures 4a and 4b:** Histogram comparing similar (left) and differing (right) distributions of categorical variables

The Kullback–Leibler (KL) divergence is a measure of how one probability distribution diverges from a second expected probability distribution, and it is often used to compare two categorical distributions. The closer the KL divergence is to 0, the more similar the distributions. In figure 4a, the KL divergence is 0.0062, suggesting a close match between the distributions. In the distribution shown in Figure 4b, the KL divergence is 0.2645, consistent with a mismatch between the distributions.

---

[3] Note that other statistical metrics, such as chi-squared test for independence, are also acceptable

Another interesting metric for comparing individual categorical variables is *support coverage*, as defined in [17], which measures how much of the variable's support (number of unique categorical values or levels) in the real data is covered in the synthetic data. Concretely, support coverage is defined as:

$$\text{Eq (2): } S\left(X_R, X_S\right) = \frac{1}{V} \sum_{v=1}^{V} \frac{\left|S^v\right|}{\left|R^v\right|}$$

Where $S^v$ represents the cardinality of synthetic variable $v$, and $R^v$ represents the cardinality of real variable $v$, and $V$ is the set of all categorical variables. A high support coverage value reflects better coverage of categorical levels in the synthetic data and thus higher fidelity.

Yet another option here is wasserstein distance[4] (also known as "earth-mover's" distance) which is often used instead of KL-divergence.

### 3.5 Pairwise Correlation

Univariate metrics do not describe interactions between variables. It is useful therefore to measure the *pairwise correlation*[5] between some or all pairs of variables in the dataset; if the pairwise correlation of the synthetic data is similar to that in the real data, then statistical fidelity is maintained.

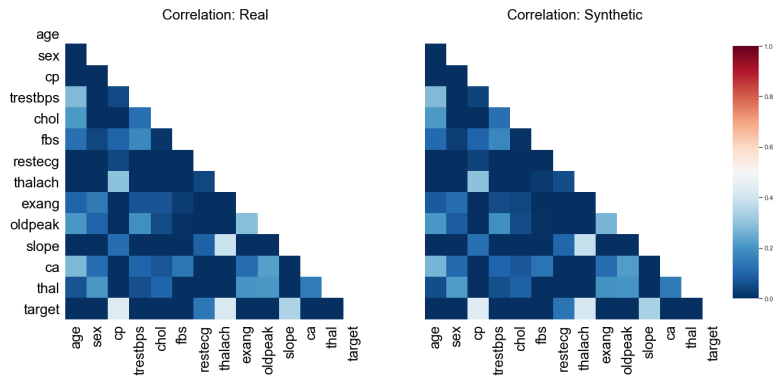

**Figure 5a:** Pairwise correlation heatmap - high fidelity

Figure 5a compares the correlation heatmap generated with the real data (left) against the same heatmap generated with the synthetic data (right) and is helpful in understanding the degree to which pairwise correlations between each pair of variables are maintained. The variables included in the heatmap are a choice of the evaluator. If the number of variables is small, all variables can be included. If the number of variables exceeds what can be understandably displayed, one can include the N most common variables in the dataset, variables of clinical importance, or use some other criteria to select these variables. It can be of interest to compare relatively *uncommon* variables in the heatmap to see how well the synthetic engine handles rare conditions or edge cases present in the real dataset.

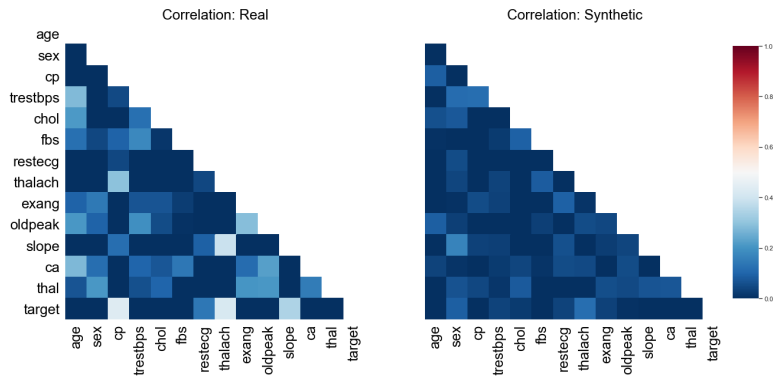

**Figure 5b:** Pairwise correlation heatmap - low fidelity

---

[4] https://en.wikipedia.org/wiki/Wasserstein_metric
[5] Herein we use Pearson correlation, but other forms of correlation (e.g. Spearman or Kendall) are equally valuable

Figure 5b demonstrates low fidelity between the real and synthetic data. Pairwise correlation can be measured quantitatively with the pairwise correlation difference (PCD), computed as the L1 or L2 norm of the difference between the correlation matrices:

$$\text{Eq (3a): } PCD_{L1} = ||Corr(real) - Corr(syn)||_1$$

$$\text{Eq (3b): } PCD_{L2} = ||Corr(real) - Corr(syn)||_2$$

Where *Corr(real)* represents the correlation matrix for the real data, and *Corr(syn)* is the same for synthetic data. PCD values of 0 or close to 0 coincide with the synthetic data being closer to the real data, and higher values mean less statistical fidelity. The highest value of PCD is 1. For example, in figure 5a, the PCD-L1 is 0.007 whereas for figure 5b, the PCD-L1 is 0.17, consistent with our expectations.

### 3.6 Multivariate Metrics

Although relatively simple to visualize and understand, both single variable and pairwise fidelity metrics lack the ability to evaluate the statistical fidelity from a perspective that takes into account all the variables and their granular linear and non-linear interactions. To address this, we propose three multivariate metrics: (1) predictive model performance (2) survival analysis (3) discriminator AUC / pMSE.

### 3.6.1 Predictive Model Performance

An effective form of fidelity validation works as follows: train two instances of a machine learning model (using a commonly accepted algorithm such as a linear regression, random forest, gradient boosted trees, or a deep neural network) - one trained with real data, and the other trained with synthetic data; then *compare the predictive performance of the models on a real-data validation set*. Using a modern predictive modeling algorithm, we can gain insight into data fidelity at the multivariate level since the models exploit non-linearities and multivariate correlations in the predictive variables.

If the target variable for the predictive model is a binary variable (classification), a common and widely used metric for measuring the performance of the predictive models is the area-under-the-curve of the receiver-operator-characteristic (ROC-AUC)[6]. Additional metrics such as accuracy, precision, recall and F1 are also informative, but to avoid too many metrics we recommend choosing only the relevant ones for the use case at hand. The ratio of the synthetic ROC-AUC and the real ROC-AUC becomes the quality metric for statistical fidelity, with higher values representing better statistical fidelity. The higher this ratio, the better the fidelity of synthetic data with the real.

If the target variable for the predictive model is a continuous variable, common and widely used metrics of performance are RMSE (root-mean-squared-error) and MAE (mean-absolute-error).

For predictive models with binary targets, we can visualize the models' performance through the ROC curve, and calculate the ratio in ROC-AUC between the model based on real data vs. the model based on synthetic data. With high fidelity synthetic data, we expect the ROC and corresponding AUC of the synthetic data to be close to each other.

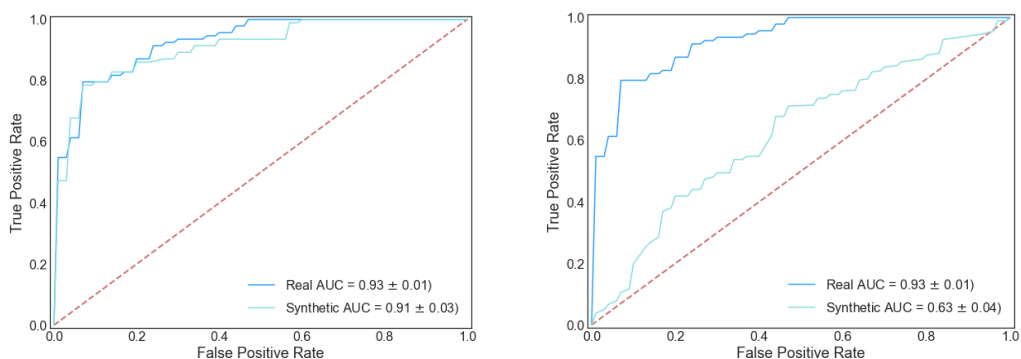

---

Note that ROC-AUC is equivalent to another statistics commonly used to summarize the ROC curve - the C-statistic.

**Figures 6a and 6b:** ROC AUC metrics comparing two predictive models with high (left) and low (right) fidelity

In figure 6a we see an example of two predictive models (generated using gradient boosted trees) where the ROC curves of the two models are nearly identical, and the AUC metrics are within a margin of error. This reflects very good fidelity of the synthetic data with the real data. In figure 6b we see the opposite case where the fidelity between the two datasets is low, resulting in the ROC curve of synthetic data closer to the red line (representing random decision).

When using predictive model performance to compare synthetic to real datasets, it is important to ensure proper data science hygiene principles:

- Use models that provide the best predictive performance. For example, in many real-world scenarios linear models underperform more advanced modeling techniques such as gradient boosted trees or random forest.
- Always set aside a separate validation set from the real dataset and evaluate performance of the predictive models on that validation set for both real and synthetic cases.
- Use hyperparameter tuning to ensure the compared models are fully optimized.
- Run the predictive model multiple times with different randomizations to ensure the outcome is not due to chance.

In practical use, the predicted variable would be one of clinical significance to the dataset being considered. For example, for general hospitalized patients, length-of-stay and readmission risk might be relevant, whereas for patients with severe COVID-19, the outcomes might be risk of admission to the ICU, risk of intubation, length of time on a ventilator, and risk of death. It is also important that there are no variables used in the prediction that leak, or anticipate, the predicted variable. For example, the lab test "type and cross for blood transfusion" would be inappropriate to use as a predictor of "likelihood of surgery in the next week."

Model performance analysis pertains to the synthetic data engine and says nothing about how useful or accurate the predictions. For example, there may be independent variables missing from the original dataset that materially impact the outcome. In predicting death due to COVID-19, unless every patient's resuscitation status is included in the input data, a predictor of death will severely underperform. Patients with "do not resuscitate" orders are much more likely to die then a patient with all other features matched, but without such an order. If a "do not resuscitate" order is absent from the features of the predictive model, the model will fail. The synthetic engine will generate a dataset equivalent to the real dataset, and the output (ROC curve) using the real and synthetic data will match, but the generative model has no way to know that a critical variable was not included.

When comparing predictive models trained on real vs synthetic data, it is useful to evaluate the "feature importance" from these models and compare the most important features between the real and synthetic datasets. Predictive models are increasingly coming under scrutiny, requiring interpretation of the model output to open up the black-box risk in such models. It follows that if a synthetic data set is an accurate representation of a real data set, the interpretation of a predictive model trained using real and using synthetic data should be very close. A leading method for computing feature importance is Shapley values (SHAP[7]) [24], which we use in our experiments and evaluation. Aside from the visualization of SHAP values, we use nDCG[8] (normalized discounted cumulative gain) to compare the ranking of feature importances and summarize the difference in a single metric; we don't expect feature rankings between real and synthetic to match exactly as small changes in training algorithms can result in minor changes in feature importance, but we do expect them to be very similar, with an nDCG value close to 1.0.

### 3.6.2 Survival Analysis

Survival analysis provides a statistical framework to analyze time-to-event outcomes. To understand whether the characteristics of time-to-event are maintained in the synthetic data, we perform Kaplan-Meier[9] analysis on the real data and then compare analysis on the synthetic data when there is a temporal aspect of the real data. In addition to the visualization of Kaplan-Meier curves, we use the p-value from the analysis as a metric to compare real data to synthetic data. Of course other types of survival analysis such as Cox Regression can be used to compare real data to synthetic data.

---

[7] An alternative is to use LIME - https://homes.cs.washington.edu/~marcotcr/blog/lime/
[8] https://en.wikipedia.org/wiki/Discounted_cumulative_gain
[9] https://www.ncbi.nlm.nih.gov/pmc/articles/PMC3059453/

### 3.6.3 Discriminator AUC

Inspired by generative adversarial networks[10], Another metric for multivariate fidelity is the "discriminator AUC". Specifically, we build a classification model trained to discriminate between the real data records and the synthetic ones. Using ROC-AUC as a measure of performance for this discriminator model, a synthetic dataset with high statistical fidelity to the real dataset will result in a ROC-AUC value for this discriminator model that is close to 0.5 (representing random decision classifier), whereas low fidelity synthetic data are reflected with ROC-AUC values closer to 1. As described above with predictive models, it's important to use a strong modeling technique and run the discriminative model multiple times with different randomizations to ensure the outcome is not due to chance.

The propensity-score mean squared error (pMSE) is a variant of ROC-AUC described in [18]. Similar to discriminator AUC, we build a discriminator model, and estimate the propensity of each record to be real or synthetic. The metric is then defined as the mean-squared-error between the propensity scores and the actual:

$$Eq\ (4):\ pMSE = \sum_{i=1}^{N} \left( p_i - c \right)$$

Where $p_i$ represents the propensity score for record i, and c is the proportion of synthetic records in the training set for the discriminator. We expect low pMSE (close to 0) for a synthetic dataset with high statistical fidelity to the real.

### 3.7 Clinical Consistency Assessment

In synthetic medical data, a final metric of fidelity for synthetic data is based on domain expertise that validates the data quality. We apply rules generated by clinicians or scientists to identify inconsistencies in the data that represent failed synthetic data generation, for example:

1. Patient is male and pregnant
2. Patient is female and has been diagnosed with prostate cancer
3. Patient is age 0-3, with adult demographics such as weight > 100 lbs, height > 5 feet, etc.
4. Patient is listed as dead, at time T but has clinical events at time T* > T

Given a library of rules, we calculate the number of records with inconsistencies as a percentage of the overall number of records.

## 4. Privacy Validation

One of the primary goals of using synthetic data is to prevent disclosure of private patient information.

We evaluate synthetic data privacy as follows:
- **Disclosure metrics**: understanding how much disclosure risk may result from access to the synthetic dataset.
- **Copy protection metrics:** demonstrating that records from the real dataset are not "copied over" to the synthetic dataset.

We assume that direct identifiers (such as first name, last name, address or social security number) are always removed from the real data before it is being used to generate synthetic data. This is often already the case for the source data, with commonly used de-identification software and/or service can be applied to achieve this initial state.

In the following we use the following notation:
- R represents the dataset of real records, and S represents the synthetic dataset
- Q is a set of quasi-identifier fields that are used by a potential attacker to match records, such as age, gender, ethnicity, zip code or similar.

### 4.1 Disclosure Metrics

There are two forms of disclosure used in classical statistical disclosure practice (not related specifically to synthetic data): *identity disclosure* and *information disclosure*. Identity disclosure is the discovery of the identity of the subject of a disclosed record, whereas information disclosure is the discovery of (additional) information about a known subject. Identity disclosure usually leads to information disclosure (once the subject of a record is identified, the record provides additional, potentially sensitive information about that individual, such as medical or psychiatric conditions, HIV status, etc.), so classical disclosure limitation aims to prevent identity disclosure, and the HIPAA Privacy Rule is explicitly framed in this way.

---

[10] https://arxiv.org/abs/1406.2661

We do note that information disclosure can happen without identity disclosure, in real datasets. For example, if a de-identified dataset contains a 3-digit ZIP code and the age and gender of patients (something generally permitted under HIPAA) and it is known from general census data that there are five individuals of a given age and gender in a certain 3-digit ZIP code, and the de-identified data contains five distinct individuals all of whom have diabetes, then the user of the de-identified data can deduce that the five individuals in that age/gender/ZIP code bucket whose identities may be found in voter registries, consumer data etc., are all diabetics, even if it is still not known how to match the de-identified records to the known identities. For handling real data, whether or not such disclosures are acceptable, needs some careful consideration. In this instance it may be undesirable, but in other cases such disclosure may only include general statistical information that can be disclosed. It is often challenging to carry out a systematic analysis of such disclosure, and this is another advantage of synthetic data.

For synthetic data, the concept of identity disclosure is somewhat ill-defined. In any meaningful application of data synthesis, where records are generated from a distribution, there is by construction no connection between a given synthetic record and any unique individual.

What is true for synthetic datasets (as well as for classically de-identified datasets) is that there can be no prohibited disclosure of identity or information about an individual, if no data associated with this individual is used in the construction of the data. Thus, a de-identified dataset will not violate the privacy of a patient not represented in the data, and a synthetic dataset will not do so if no information associated with the individual was used to train the synthetic algorithm. The only information about an individual that could be leaked from a synthetic dataset created without the use of that patient's true data is general statistical information that we do not aim to conceal. For example, if a synthetic dataset reveals that all patients who have an appendectomy performed have been diagnosed with appendicitis, and the user of that synthetic dataset concludes that their work colleague who mentioned having had their appendix removed must have had such a diagnosis, this would not be an illegitimate information disclosure.

Thus, to understand disclosure risk for synthetic data we focus on *membership inference* and *attribute inference* attacks. In a membership inference attack the adversary aims to identify the participation of a known patient record in the training of the synthetic generation algorithm, and in an attribute inference attack the adversary attempts to infer values of one or more sensitive attributes from the synthetic dataset.

### 4.1.1 Membership Inference Test
Membership inference attacks seek to infer membership of a patient record in the real dataset from which the synthetic data was generated. For example, if the training set used to train a generative model consists of HIV positive patient records (e.g., the HIV status is not included as a field, but it is a clinical study of HIV patients), then inferring whether a patient record was included in that training will reveal that that patient is HIV positive.

We now describe a validation test for membership inference on synthetic data. Following [21] let $R$ be a (large) training set of real patient data, and $r \in R$ a patient record. For simplicity of presentation, we assume that R contains one unique record per patient, though this limitation can be easily removed. We split the training data randomly into two disjoint subsets of equal size, $R = R_1 \cup R_2$. We train the generative model on $R_1$ and generate a synthetic data set $S_1$. We utilize a reasonable measure of distance between the source patient record $r$ and any synthetic record $s$, such as the one described in 3.1.

Even though there is no well-defined concept of the "identity of a synthetic record", we can design a well-defined test for "membership inference". Concretely, we consider a hypothetical adversary which has access to a subset of records in R which we denote $R_3$ (note that records in $R_3$ may belong to either $R_1$ and $R_2$), and that attempts a membership inference "attack" as follows:

- Given a patient record $r \in R_3$ , and the disclosed synthetic dataset $S_1$, the adversary identifies the closest record $s \in S_1$ with distance $d(r, s)$ .

- The adversary determines that $r$ is part of the training set of $S_1$ if $d(r, s)$ is lower than some threshold chosen by the adversary (in our simulation we use the minimum Hamming distance that provides a match H, H+1, H+2, H+3, and H+5).

We want to evaluate the success rate of such an adversarial strategy. For each record $r \in R_3$ we know if it actually belongs to $R_1$ or $R_2$, and can then determine whether the adversary's decision constitutes a true positive, true negative, false positive or false negative, and measure the confusion matrix. We then compute precision[11] and recall and plot them as a function of the % of records in R that are present in $R_3$.

Figure 7a describes in a visual form the approach for the membership inference attack:

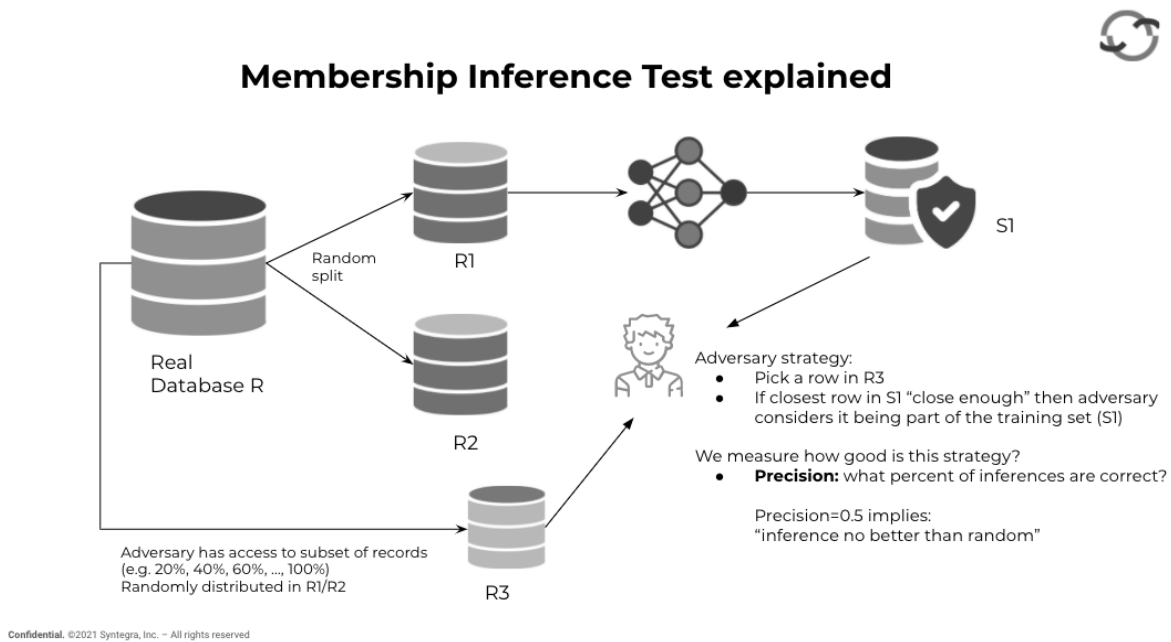

**Figure 7a**: membership inference diagram

Precision represents the number of correct decisions the adversary has made; since we randomly split R into $R_1$ and $R_2$ with equal sizes, the baseline precision is 0.5 (corresponding to random choice), and any value above that reflects increasing levels of disclosure risk (or increased risk of success for the membership inference attack). Recall (aka sensitivity) represents the percent of records known to an attacker that can be found in the training set. Clearly as the Hamming distance threshold increases (and recall increases), the attacker identifies more and more such records as belonging to the training set (although that doesn't mean such identification is correct as reflected in the precision). This is demonstrated in figure 7b below:

---

[11] https://en.wikipedia.org/wiki/Precision_and_recall

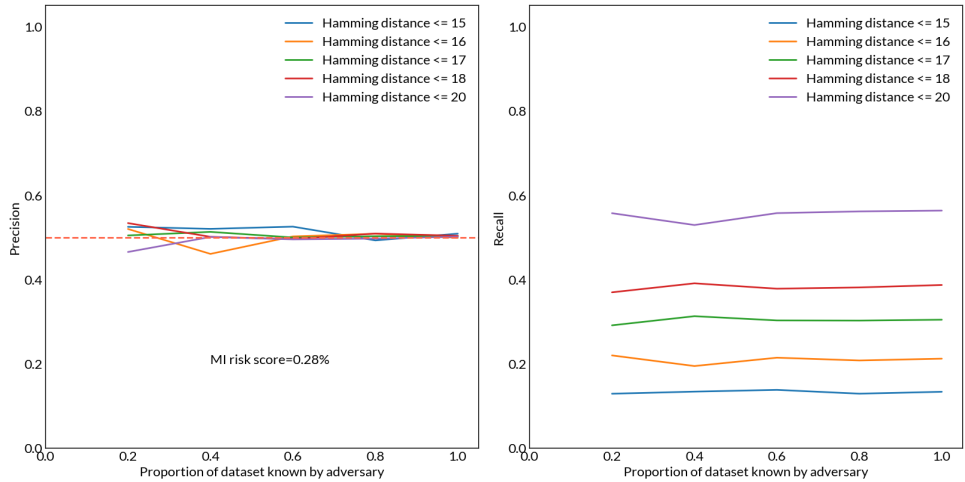

**Figure 7b**: membership inference metrics

As can be seen in this example, the precision of membership inference attack is very close to the baseline of 0.5 and thus represents minimal risk of disclosure. From the example in Figure 7b, it is clear that no matter the threshold chosen by the adversary (Hamming distance threshold chosen, reflected in the recall), the likelihood of making a correct membership inference metric is very close to "random guess".

As a summary metric, we look at the precision of membership inference at the first threshold reflecting recall level of 50% or higher (in figure 7b above hamming distance <= 20), normalized to a value between 0 and 1

$$MI\ Risk\ Score = (precision - 0.5) * 2$$

We consider MI Risk score < 0.2 (20%) as very low risk of disclosure due to membership inference.

### 4.1.2 File Membership Hypothesis Test

We now present a slight variant of the membership inference test, where we perform a random membership inference test N times, and use statistical hypothesis testing to gain insight into possible success of a membership inference attack. In a similar manner to 4.1.1, we split the training data randomly into two disjoint subsets of equal size, $R = R_1 \cup R_2$. We then train the generative model on $R_1$ to generate a synthetic data set $S_1$, and train the generative model on $R_2$ to generate a synthetic data set $S_2$. For any patient record $r \in R$, we find the closest records $s_1 \in S_1$ and $s_2 \in S_2$. If $d(r, s_1) < d(r, s_2)$ we say $r$ is in $S_1$, and otherwise $r$ is in $S_2$. We deem this determination to be correct if $r \in S_1$ and false otherwise.

Running this process N times, each time performing the randomization using a different random seed, we can perform hypothesis testing. Let $P$ be the probability that our test is "correct", or experimentally the rate of "correct" answers over a large sample. We then make the null hypothesis H0: $P \neq 1/2$. Using standard Central Limit Theorem methods we can then aim to reject H0. If we succeed in doing so, this is a powerful indication of privacy. It states that we are unable to guess with any greater accuracy than a coin toss whether a patient's records were even present in the data used to train a model. This precludes any identity or information disclosure.

We consider the synthetic dataset S to represent very low disclosure risk if we are able to reject the null hypothesis (p-value<0.05).

### 4.1.3 Attribute Inference Test

Following [21] and [27] we split the dataset variables into two subset: (1) Q (quasi-identifiers, e.g. age, gender or race) which is a subset of variables in the dataset that the attacker may possess about real patients and (2) the rest of the variables, including some sensitive variables the attacker may try to infer.

For a specific person, the attacker obtains a given set of values of the quasi-identifier variables Q, and filters the synthetic dataset to those records that match the quasi-identifiers ($S_q$); by default an exact match is performed to establish $S_q$ but an approximate match can also be performed whereby a row in S is matched for each field in Q within a certain range of values around the exact values obtained by the attacker. Now with $S_q$ the attacker infers the value of the sensitive variable t for that person as follows:

1. If t is categorical, the attacker selects the mode of $S_q(t)$ as the inferred value
2. If t is numeric, the attacker computes the median of $S_q(t)$ as the inferred value

Alternatively the attacker can train a machine-learning model (such as a random forest or gradient boosted trees model) using the synthetic dataset S as a training set (where the quasi-identifiers Q are used as predictors and the sensitive variable t is used as the target) and then predict the value of t using this model.

In our evaluation we follow [27] and estimate the disclosure risk for a sensitive variable t as follows:

$$\text{Eq (6): } disclosure\ risk(t) = \frac{1}{n_s} \sum_{s=1}^{n_s} \left( \frac{1}{f_s} \cdot \lambda_s' \cdot I_s \cdot R_s \right)$$

Where

- $n_s$ is the number of records in the real dataset
- $f_s$ is the equivalence class size in the real sample for particular records s in that real sample
- $\lambda_s'$ is the adjustment factor due to errors in matching; in [27] the authors use $\lambda_s' = \frac{1 + (0.23x(1-0.0426)^k}{2}$ where k is the number of quasi-identifier fields (see that paper for more details).
- $I_s$ is an indicator variable that takes a value of 1 if the record s in the real data has a matching record in the synthetic data, and 0 otherwise
- $R_s$ is an indicator variable that takes a value of 1 if the adversary would learn something new from matching the records s to the synthetic data

This disclosure risk represents the percentage of records which, for this sensitive variable t, represent high risk of disclosure. We consider a disclosure risk of 0.05 (5%) or lower to be very low and acceptable for practical purposes.

Our methodology closely follows [27], however with a few improvements:

- For each variable <t> not included in the Quasi-identifiers, we compute and report the disclosure risk for this variable as specified above. We consider the overall disclosure risk for a patient record as the maximum disclosure risk for that patient record across all variables (as opposed to [27] where they consider a patient record at risk if L% of variables are at risk of disclosure).
- We use Jenks Natural Breaks to bin numeric values into K bins (as a more robust approach to using k-means clustering suggested in [27]); we estimate the optimal value of K by calculating goodness-of-fit and picking the first K where it's above 0.8.
- Missing values are considered lower risk, if disclosed. By default, we reduce the risk by 50%, and this can be configured manually as part of risk assessment. This weighting is associated with Rs, reflecting the fact that a missing value discloses less "new information" if missing.
- We note that the definition of $\lambda_s'$ is based on field experience and other peer reviewed studies as described in [27] and they provide a strong baseline; we also recommend and implement two other variants that represents stronger privacy guarantees: (1) a more conservative estimate of 80% match rate and 0% error rate (2) a no-error estimate of 100% match rate and 0% error rate.

We report for each target variable the disclosure risk as computed above, and exemplified in table 2:

| Target Variable | COVID+ | HIV+ | Diabetic | Total |
|---|---|---|---|---|
| Reference paper | 0.00234 | 0.0234 | 0.000012 | 0.00243 |
| Conservative | 0.00534 | 0.0266 | 0.00003 | 0.00331 |
| No Errors | 0.00662 | 0.0345 | 0.00043 | 0.00532 |

**Table 2**: Example attribute inference results for three sensitive variables: COVID+, HIV+ and diabetic. We report three variants of the disclosure risk: using the error estimate used in [27], a more conservative estimate, and the most conservative (no errors) estimate.

Although not reported in our results here, we note a possible future enhancement where a "sensitivity weight" is associated with each variable, and the disclosure risk is properly adjusted accordingly. For example, any disclosure associated with a variable about COVID positive status or HIV positive status is more sensitive than disclosure of a patient's height or age.

## 4.2 Copy Protection Metrics

An important question when looking at high fidelity synthetic data that is derived from real data is "how do I know that the synthetic data is not a simple copy or minor perturbation of the original real data", resulting in high risk of disclosure?

To address this issue, we propose two methods:
- Distance-to-closest-record (or DCR)
- Exposure to unintended memorization

### 4.2.1 Distance to Closest Record - DCR

For any given real patient record $r \in R$ we define DCR($r$) as the distance between $r$ and that record in the synthetic dataset $s \in S$ that is closest to it. Assuming an appropriate distance metric d(r,s) as discussed in 3.1, to compute the DCR for a given record $r$ in the real dataset we take the minimal distance to all candidate records in the synthetic dataset $s_j$ and obtain that row's DCR value.

$$Eq\ (7): DCR(r) = min_j\ d(r, s_j)$$

The intuition here is that DCR provides us with an understanding of the distance between records in the real and synthetic datasets, and any situation where exact copies or simple perturbations of the real records that exist in the synthetic dataset will be easily exposed by the DCR metric.

Figure 8a demonstrates a privacy-preserving synthetic dataset where DCR values are "far" from 0, whereas figure 8b demonstrates a situation where some synthetic records exactly match the real records, while some other do not, which represents a high risk of unintended copying of at least some of the records (those with DCR=0).

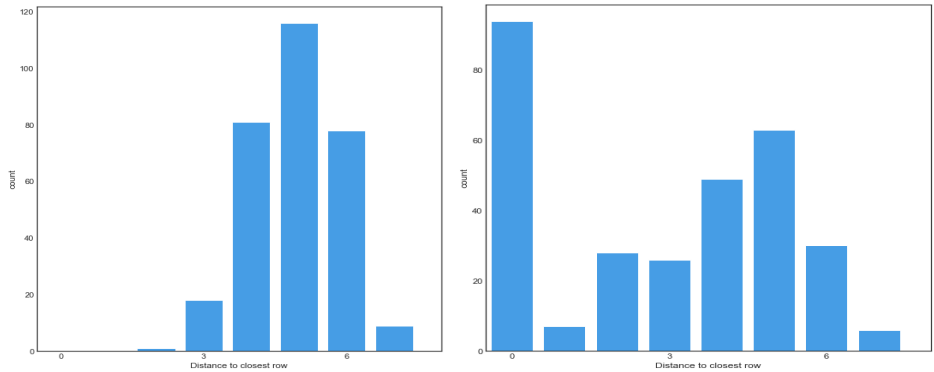

**Figures 8a and 8b:** DCR distribution example: privacy preserving synthetic dataset (left) and synthetic data containing many copies of the original (right).

For privacy-preserving synthetic data we expect DCR between the real dataset and the synthetic dataset (real-synthetic) look like figure 8a where the lowest DCR value is far from 0, and not like figure 8b where we see a large number of exact copies of real records that have an exact match in the synthetic dataset.

We note, however, that although visualizing the DCR distribution in this way provides useful insights into privacy preservation, DCR=0 in itself doesn't necessarily map to high disclosure risk - in some datasets the "space" spanned by the variables in scope is relatively small. As an extreme example, if we only have 3 variables: gender (with values male/female), Race (with values White or Non-White) and COVID-positive (with values 0 or 1), then there are exactly 8 possible records and any dataset with more than 8 records would have at least 2 records that are exactly the same (DCR=0). Therefore, we propose the following metric to estimate disclosure risk for records with DCR=0:

1. We call $R_0$ the subset of real records records with DCR=0.
2. For all records in $R_0$, we calculate all the possible equivalence classes and consider those with an equivalence class (in R) size of 5 or less as "high risk rows" - $R_{HR}$.
3. We consider a synthetic dataset risk disclosure low when the percent of high risk rows out of the overall dataset - computed as $|R_{HR}|$ / |R| - is less than 1%, following industry best practices in re-identification analysis.

### 4.2.2 Exposure

Our second copy protection test is specific to synthetic data engines that can associate a probability to any given record (or sample) $P(record)$, representing how likely that record is to occur in the synthetic dataset; as has been recently demonstrated in [20] and [26], this is useful in order to assess whether or not a synthetic data generator might be unintentionally memorizing the input data.

Following [20] we propose to measure the extent of unintended memorization by injecting "canary samples" (values far outside the distribution of real data) into the training set and measuring log-perplexity (defined as $log(P(sample))$) and *exposure* in the synthetic dataset. Specifically, for a given synthetic dataset generated using a synthetic engine, we define for a given canary record c injected into the training data the exposure as:

$$Eq\ (8): exposure(c) = log_2(|C|) - log_2(rank(c))$$

Where rank(c) is the rank of the chosen canary amongst all possible canaries and C is the space of all possible canaries. Exposure is a real value ranging between 0 and $log_2(|C|)$; its maximum can be achieved only by the most-likely, top-ranked canary; conversely, its minimum of 0 is the least likely; across possibly inserted canaries, the median exposure is 1.

Thus, our proposed metric works as follows:
- Insert N randomly chosen canaries into the training set R, resulting in R' (original R + the canaries)
- Generate a synthetic dataset S' from R'
- Calculate the average exposure over the N canaries in S'

By estimating the size of C, we can calculate the "extraction threshold" as $log_2(|C|)$ and measure the exposure level against that threshold. If the actual exposure is lower than the threshold, we determine that the risk of unintended memorization is low, whereas if the exposure is higher than the threshold then the risk is high.

Note that this method requires access to the synthetic data generation engine (as opposed to just access to a given synthetic dataset), as the canaries need to be included in the training dataset.

## 5. Experimental Results

We evaluate our fidelity and safety metrics with four datasets, which we abbreviate as: DIG, NIS, TEXAS, and BREAST. For each dataset, we generate synthetic data using Syntegra's engine, and compute the fidelity and privacy validation metrics. Note that for some metrics, like ML modeling ROC-AUC or discriminator AUC we actually ran this evaluation 5 times, each time generating synthetic data using a different randomization seed, and measured the metrics in each case, using mean and std-deviation to report the average and standard-deviation of each metric.

### 5.1 Datasets

The Digitalis Investigation Group (DIG) study [31] was a randomized, double-blind, multicenter trial with more than 300 centers in the United States and Canada. The purpose of the trial was to examine the safety and efficacy of digoxin in treating patients with congestive heart failure (CHF), in sinus rhythm, with an ejection fraction ≤ 0.45. Endpoints were mortality from any cause (the primary endpoint) and hospitalization for heart failure (the secondary end point) over a three-to-five-year period. We downloaded openly available, participant-level data from the NIH/NHLBI BioLINCC data repository. This dataset includes 6,800 patient records and 71 variables, and we generated a dataset of 20,000 synthetic patient records.

The National Inpatient Sample database[12] (NIS) is a patient-level administrative claims database that represents approximately 20% of discharges from US community hospitals. The NIS includes data on patient demographics, primary and secondary diagnoses, medical comorbidities, surgical procedures, length of stay, and discharge disposition. We utilized a version of this dataset with 116,009 patient records and 31 variables and generated a dataset with 200,000 synthetic patient records.

The TEXAS Hospital Discharge dataset[13] is a large public use data file provided by the Texas Department of State Health Services. The dataset we used consists of 50,000 records uniformly sampled from a pre-processed data file that contains records from 2013, similar to how it is used in [23], retaining the same 18 data attributes (of which 11 are categorical and 7 continuous attributes). We generated a dataset with 100,000 synthetic patient records.

The National Cancer Institute Surveillance, Epidemiology, and End Results Program (SEER) collects cancer incidence data from population-based cancer registries covering approximately 35 percent of the U.S. population. The SEER registries collect data on patient demographics, primary tumor site, tumor morphology, stage at diagnosis, and first course of treatment, and they follow up with patients for vital status. We used the portion of the SEER dataset comprised of breast cancer patient records, taken from the SEER Incidence database[14], and which consists of 1,072,173 patient records with 117 variables including demographics, cancer diagnosis and classification, and other related fields. We generated a dataset with 1,500,000 synthetic patient records.

Table 3 below summarizes the characteristics of all the datasets included in our study:

| Dataset | Rows in original data | Columns | Rows generated |
|---|---|---|---|
| DIG | 6,800 | 71 | 20,000 |
| NIS | 116,009 | 31 | 200,000 |
| TEXAS | 50,000 | 18 | 100,000 |
| BREAST | 1,072,173 | 117 | 1,500,000 |

**Table 3**: dataset characteristics summary

## 5.2 Results
In this section we describe our experimental results. In most cases we compare the original dataset to the full synthetic dataset (which is often generated with a larger number of records), except when specifically mentioned.

### 5.2.1 Statistical Fidelity
First, we look at population statistics for each dataset, comparing real to synthetic for selected variables in each dataset. We randomly sample the synthetic dataset to the same size as the original. The results are demonstrated in tables 4a-4d below:

Numeric variables

| variable | Real | | Synthetic | |
|---|---|---|---|---|
| | mean | SD | mean | SD |
| AGE | 63.5 | 10.9 | 63.2 | 10.8 |
| BMI | 27.1 | 5.2 | 27.3 | 5.2 |
| HEART_RATE | 78.8 | 12.7 | 78.6 | 12.6 |
| EJF_PER | 28.5 | 8.8 | 28.6 | 8.9 |

Categorical variables

| variable | Val | Real | | Synthetic | |
|---|---|---|---|---|---|
| | | n | % | n | % |
| SEX | Female | 1,519 | 22.3 | 1,476 | 21.7 |
| | Male | 5,281 | 77.7 | 5,324 | 78.3 |
| RACE | NonWhite | 991 | 14.6 | 1,018 | 15.0 |
| | White | 5,809 | 85.4 | 5,782 | 85.0 |
| FUNCTCLS | Class I | 907 | 13.3 | 913 | 13.4 |
| | Class II | 3,664 | 53.9 | 3,653 | 53.7 |
| | Class III | 2,081 | 30.6 | 2,081 | 30.6 |
| | Class IV | 142 | 2.1 | 148 | 2.2 |

**Table 4a:** DIG - population statistics - example numeric and categorical variables

Numeric variables

| variable | Real | | Synthetic | |
|---|---|---|---|---|
| | mean | SD | mean | SD |
| age | 66.6 | 13.7 | 66.4 | 13.8 |
| Length of stay | 7.5 | 9.2 | 7.4 | 9.8 |

Categorical variables

| variable | Val | Real | | Synthetic | |
|---|---|---|---|---|---|
| | | n | % | n | % |
| Major amputation | No | 105,251 | 90.7 | 105,053 | 90.6 |

---

| | | | | | |
|---|---|---|---|---|---|
| | Yes | 10,757 | 9.3 | 10,955 | 9.4 |
| Sex | male | 67,114 | 57.9 | 66,131 | 57.0 |
| | female | 48,892 | 42.1 | 49,876 | 43.0 |

**Table 4b:** NIS - population statistics - example numeric and categorical variables

Numeric variables

| variable | Real | | Synthetic | |
|---|---|---|---|---|
| | mean | SD | mean | SD |
| Total charges ($) | 43,473 | 79,960 | 45,155 | 83575 |
| Total charges accomod ($) | 9,572 | 23,859. | 9,653 | 23814 |
| Length of stay | 5.2 | 10.4 | 5.1 | 9.7 |

Categorical variables

| variable | Val | Real | | Synthetic | |
|---|---|---|---|---|---|
| | | n | % | n | % |
| Admit weekday | 1 | 8,659 | 17.3 | 9,189 | 18.4 |
| | 2 | 8,604 | 17.2 | 7,918 | 15.8 |
| | 3 | 8,053 | 16.1 | 7,749 | 15.5 |
| | 4 | 7,963 | 15.9 | 8,317 | 16.6 |
| | 5 | 7,400 | 14.8 | 7,353 | 14.7 |
| | 6 | 4,717 | 9.4 | 4,874 | 9.7 |
| | 7 | 4,604 | 9.2 | 4,600 | 9.2 |
| RACE | American indian/eskimo | 129 | 0.3 | 108 | 0.2 |
| | Asian or pacific islander | 1,026 | 2.1 | 980 | 2.0 |
| | Black | 6,585 | 13.2 | 6,321 | 12.6 |
| | White | 31,389 | 62.8 | 31,118 | 62.2 |
| | Other | 10,865 | 21.7 | 11,472 | 22.9 |

**Table 4c:** Texas - population statistics - example numeric and categorical variables

Numeric variables

| variable | Real | | Synthetic | |
|---|---|---|---|---|
| | mean | SD | mean | SD |
| Tumor size | 96.6 | 258.3 | 88.7 | 246.3 |
| # in-situ malignant | 1.4 | 0.6 | 1.3 | 0.6 |

Categorical variables

| variable | Val | Real | | Synthetic | |
|---|---|---|---|---|---|
| | | n | % | n | % |
| sex | Female | 99,286 | 99.3 | 99,361 | 99.4 |
| | Male | 714 | 0.7 | 639 | 0.6 |
| grade | Grade I | 20,060 | 22.1 | 20,723 | 22.4 |
| | Grade II | 39,519 | 43.5 | 40,378 | 43.7 |
| | Grade III | 30,462 | 33.5 | 30,428 | 33.0 |
| | Grade IV | 901 | 1.0 | 801 | 0.9 |

**Table 4d:** BREAST - population statistics - example numeric and categorical variables (sample size reduced from 1,500,000 to 100,000).

We ran our univariate analysis across all variables in each dataset, comparing marginal distributions between real and synthetic. Given limited space we will display here a few selected variables for each dataset. Note that some variables have non-standard distributions causing some visualizations with histograms to have strange artifacts, however the match between real and synthetic is still maintained.

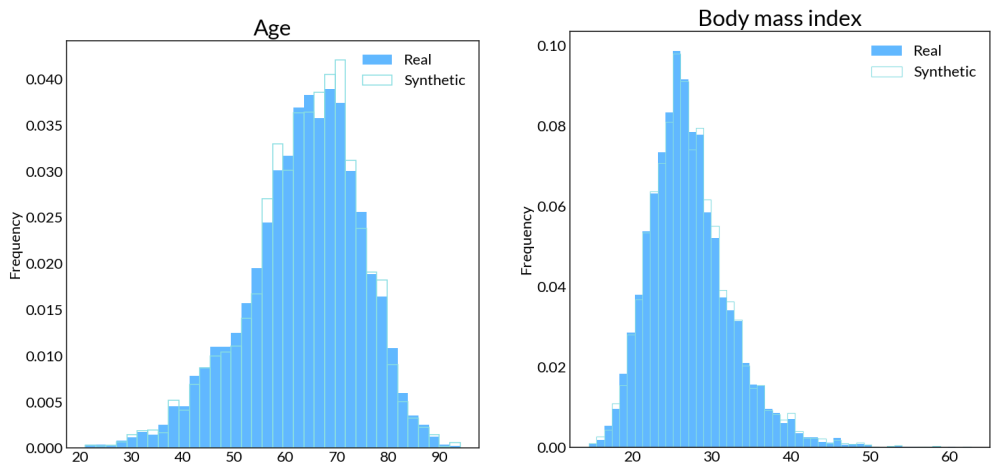

**Figure 9a:** DIG - real vs synthetic - marginal distributions for numeric variables: AGE and BMI

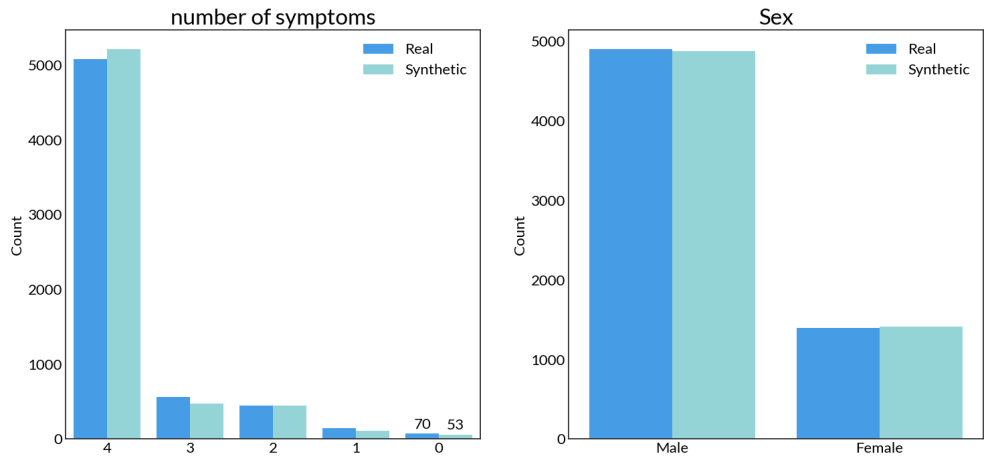

**Figure 9b:** DIG - real vs synthetic - marginal distributions for categorical variables: no. of symptoms and sex

Due to limited space, we are not showing visualizations for all variables; however, for the DIG dataset table 5 lists most of the relevant numerical metrics:

| Variable | KL-Div |
|---|---|
| Ace inhibitor | 0.00030 |
| Current angina | 0.00324 |
| Death or hosp from WHF | 0.00109 |
| Death? | 0.00266 |
| Digoxin within past week | 0.00086 |
| Dose of digoxin or placebo | 0.00165 |
| Dyspnea on exertion | 0.00547 |
| Dyspnea at rest | 0.00522 |
| EF method | 0.00026 |
| Elevated jugular venus pressure | 0.00243 |
| Etiology of CHF | 0.00335 |
| History of diabetes | 0.00443 |
| History of hypertension | 0.00139 |
| Hosp: any hospitalization | 0.00508 |
| Hosp: cardiovascular disease | 0.00377 |
| Hosp: coronary revascularization | 0.00041 |
| Hosp: digoxin toxicity | 0.00000 |
| Hosp: MI | 0.00218 |
| Hosp: other cardiovascular event | 0.00158 |
| Hosp: respiratory infection | 0.00143 |
| Hosp: stroke | 0.00086 |
| Hosp: supraventricular arrhythmia | 0.00166 |
| Hosp: unstable angina | 0.00185 |
| Hosp: ventricular arrhythmia | 0.00039 |
| Hosp: worsening heart failure | 0.00052 |
| Number of hospitalizations | 0.01114 |
| Number of symptoms | 0.00326 |
| NYHA Functional class | 0.00012 |
| Peripheral edema | 0.00326 |
| Previous MI | 0.00054 |
| Race | 0.00041 |
| Reason for death | 0.00433 |
| Recommended digoxin dose | 0.00167 |
| S3 gallop | 0.00302 |
| Sex | 0.00023 |
| Treatment | 0.00001 |

| Variable | Wass. Dist. | KS-Val | KS-p |
|---|---|---|---|
| Age | 0.0224 | 0.0127 | 0.6889 |
| Body mass index | 0.0220 | 0.0136 | 0.5962 |
| Chest x-ray (CT ratio) | 0.1014 | 0.0511 | <0.0001 |
| Days until 1st hosp | 0.1453 | 0.0648 | <0.0001 |
| Days until CREV | 0.1071 | 0.0620 | <0.0001 |
| Days until CVD | 0.1223 | 0.0623 | <0.0001 |
| Days until death | 0.0988 | 0.0585 | <0.0001 |
| Days until DIG | 0.1016 | 0.0602 | <0.0001 |
| Days until DWHF | 0.0815 | 0.0540 | <0.0001 |
| Days until MI | 0.1051 | 0.0612 | <0.0001 |
| Days until OCVD | 0.1184 | 0.0666 | <0.0001 |
| Days until OTH | 0.1382 | 0.0729 | <0.0001 |
| Days until RINF | 0.1091 | 0.0637 | <0.0001 |
| Days until STRK | 0.1065 | 0.0613 | <0.0001 |
| Days until SVA | 0.1149 | 0.0650 | <0.0001 |
| Days until UANG | 0.1193 | 0.0650 | <0.0001 |
| Days until VENA | 0.1024 | 0.0601 | <0.0001 |
| Days until WHF | 0.0814 | 0.0540 | <0.0001 |
| DBP (mmHg) | 0.0369 | 0.0158 | 0.4063 |
| Duration of CHF (months) | 0.0181 | 0.0115 | 0.7925 |
| EF (percent) | 0.0456 | 0.0287 | 0.0109 |
| Heart Rate (beats/min) | 0.0445 | 0.0318 | 0.0033 |
| SBP (mmHg) | 0.0693 | 0.0389 | 0.0001 |
| Serum Cr (mg/dL) | 0.0957 | 0.0531 | <0.0001 |
| Serum K level | 0.0193 | 0.0178 | 0.3346 |

**Table 5:** full univariate metrics for DIG dataset

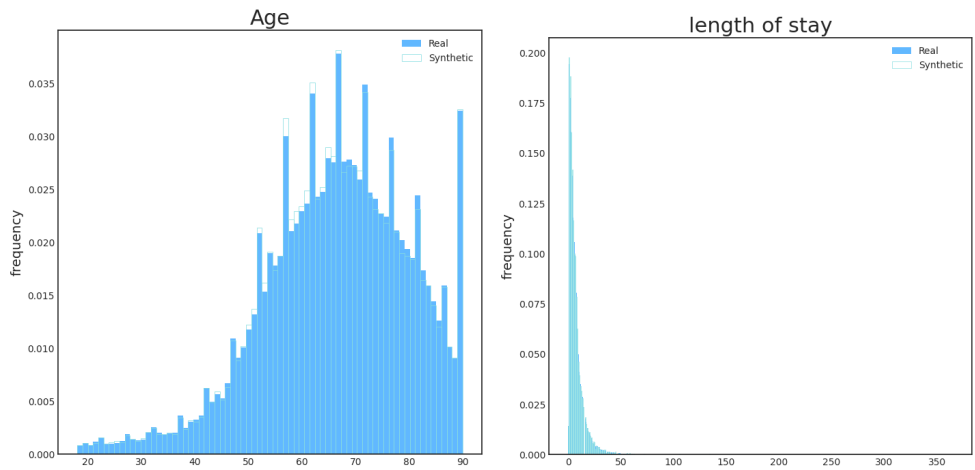

**Figure 9c:** NIS - real vs synthetic - marginal distributions for numeric variables: Age and length of stay

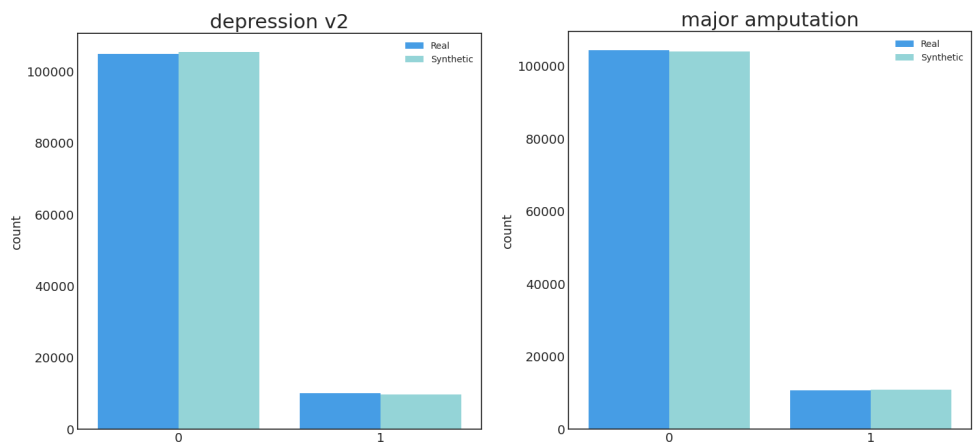

**Figure 9d:** NIS - real vs synthetic - marginal distributions for categorical variables: depression and amputation

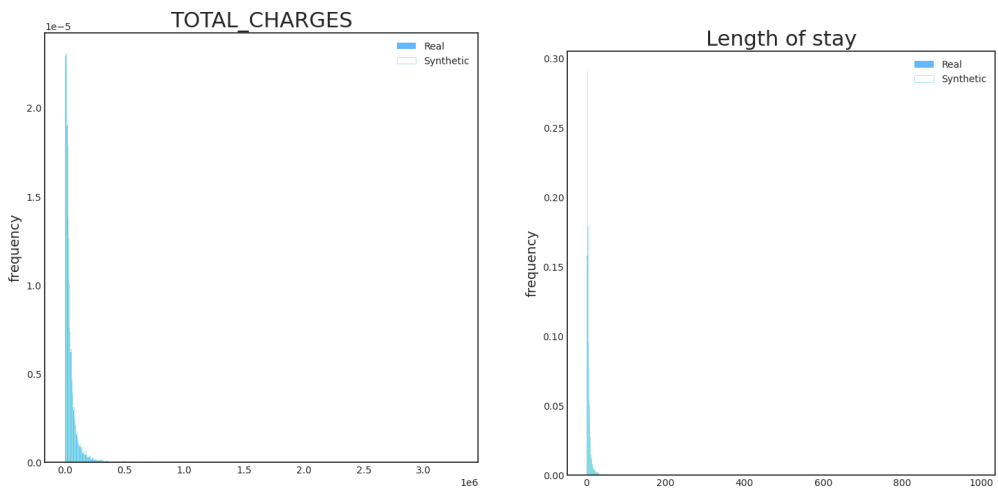

**Figure 9e:** TEXAS - real vs synthetic - marginal distributions for numeric variables

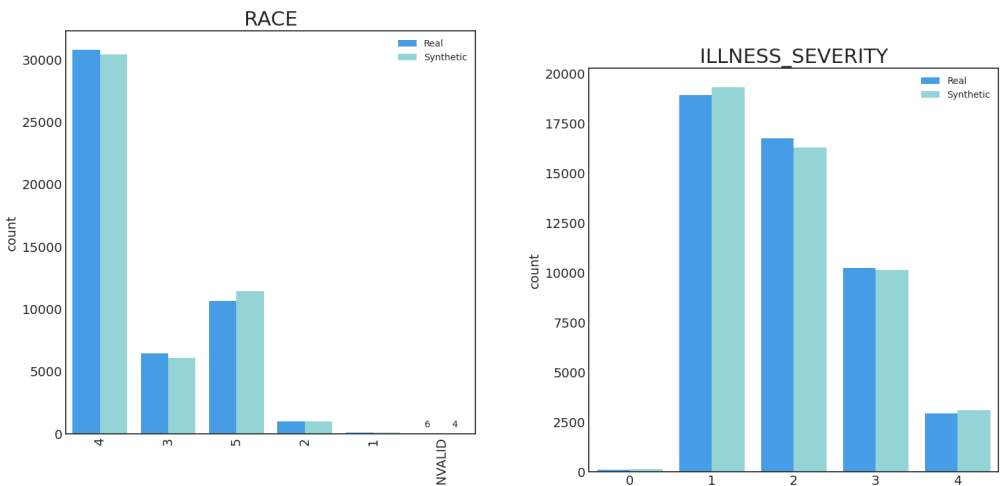

**Figure 9f:** TEXAS - real vs synthetic - marginal distributions for categorical variables: Race and Severity of Illness

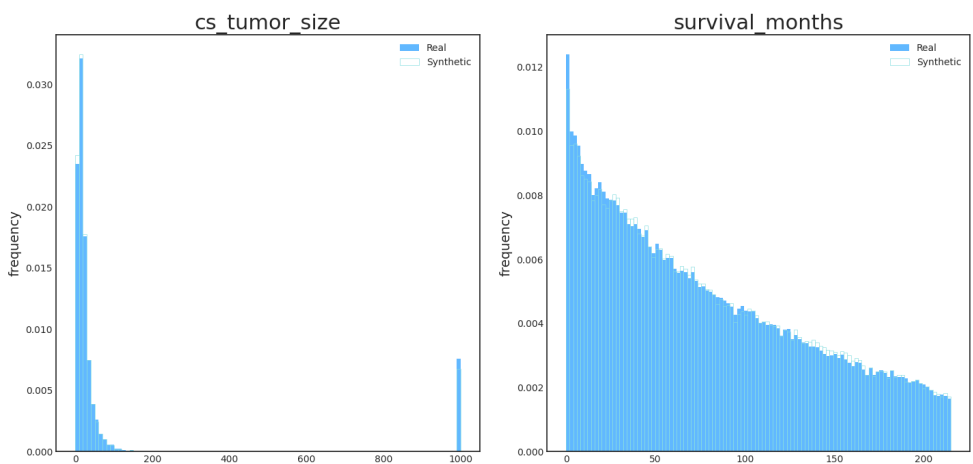

**Figure 9g:** BREAST - real vs synthetic - marginal distributions for numerical: cs tumor size and survival in months

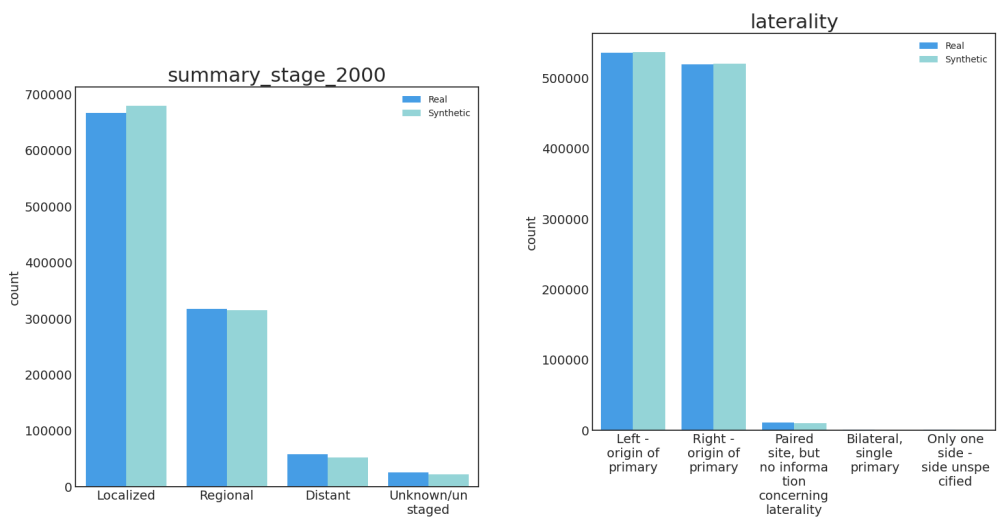

**Figure 9h:** BREAST - real vs synthetic - marginal distributions for categorical variables: summary stage and laterality

In table 6 we show the marginal distribution match metrics; recall that for numeric variables we use the KS-Statistic and for categorical variables we use the KL-divergence, and in both cases a value close to 0.0 reflects a good match in distributions:

| Dataset | Variable | Metric | Value |
|---------|----------|--------|-------|
| DIG | AGE | KS-Stat | 0.0082 |
| DIG | BMI | KS-Stat | 0.0156 |
| DIG | Num of Symptoms | KL-Div | 0.0051 |
| DIG | SEX | KL-Div | 0.0004 |
| NIS | Age | KS-Stat | 0.0095 |
| NIS | Length of stay | KS-Stat | 0.0174 |
| NIS | Depression | KL-Div | 0.0001 |
| NIS | Major Amputation | KL-Div | <0.0001 |
| TEXAS | Total charges | KS-Stat | 0.0160 |
| TEXAS | Length of stay | KS-Stat | 0.0209 |
| TEXAS | Race | KL-Div | 0.0008 |
| TEXAS | Severity of illness | KL-Div | 0.0003 |
| BREAST | CS tumor size | KS-Stat | 0.0128 |
| BREAST | Survival months | KS-Stat | 0.0082 |
| BREAST | Summary stage | KL-Div | 0.0006 |
| BREAST | laterality | KL-Div | 0.0002 |

**Table 6:** univariate distribution metrics

Next, we compute pairwise correlations between variables and plot that as a heatmap, where the strength of the correlation is color-coded with dark blue representing low correlation and red representing strong correlation:

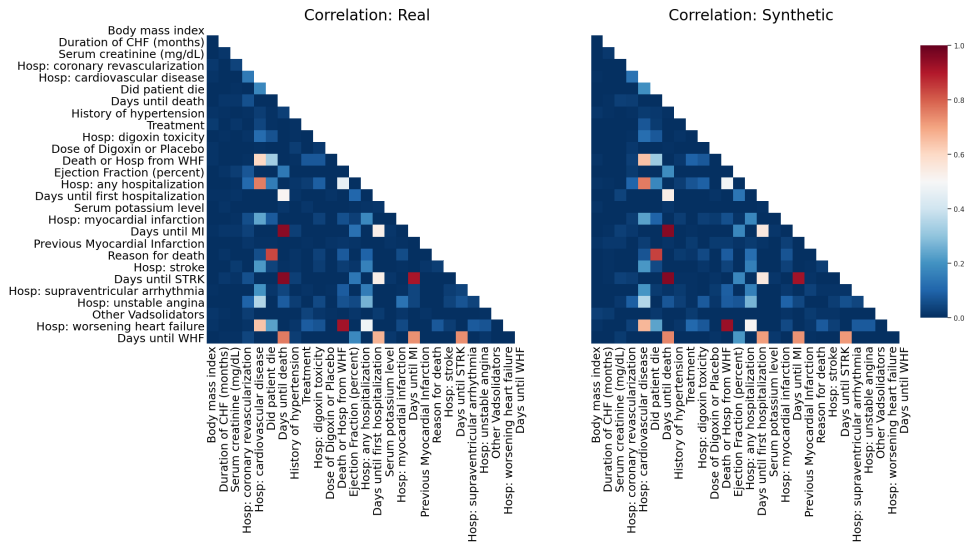

**Figure 10a:** real vs synthetic - DIG - pairwise correlations; PCD-L1 = 0.0153

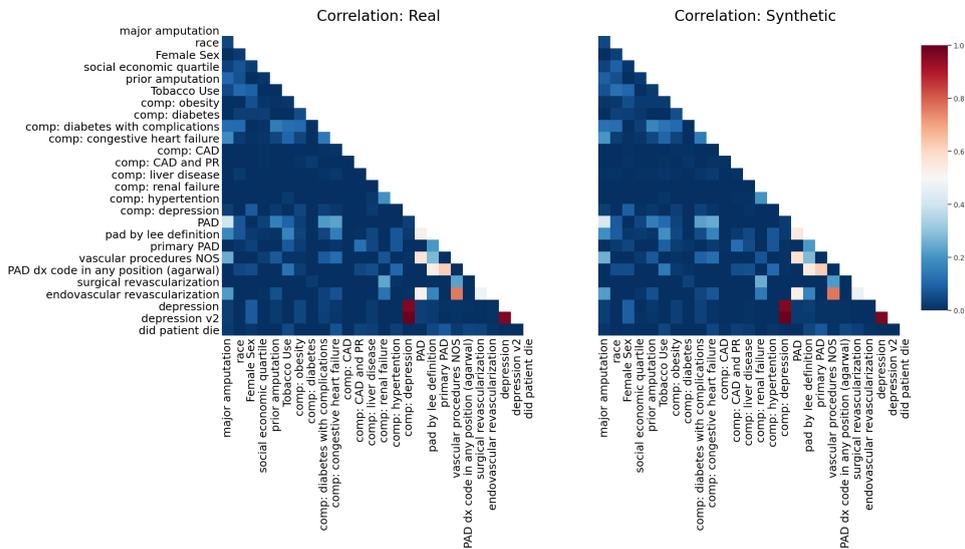

**Figure 10b:** real vs synthetic - NIS - pairwise correlations; PCD-L1 = 0.0071

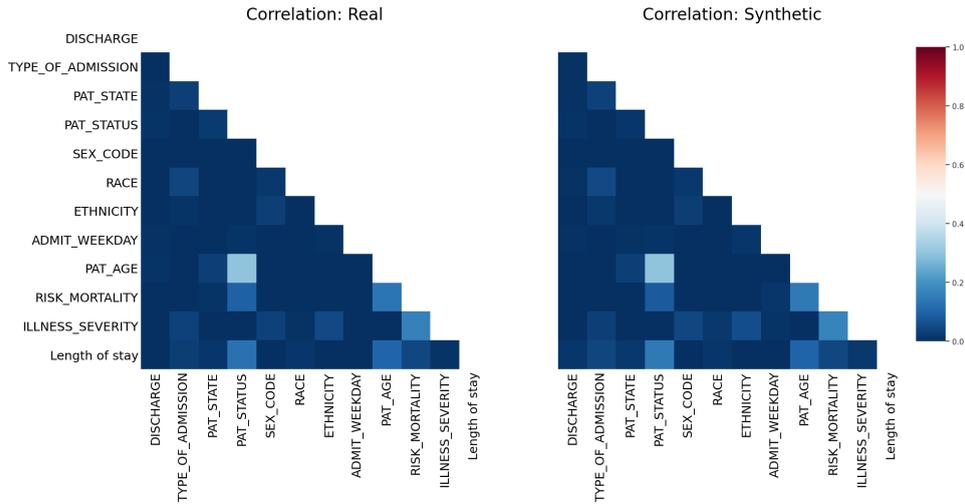

**Figure 10c:** real vs synthetic - TEXAS - pairwise correlations; PCD-L1 = 0.0056

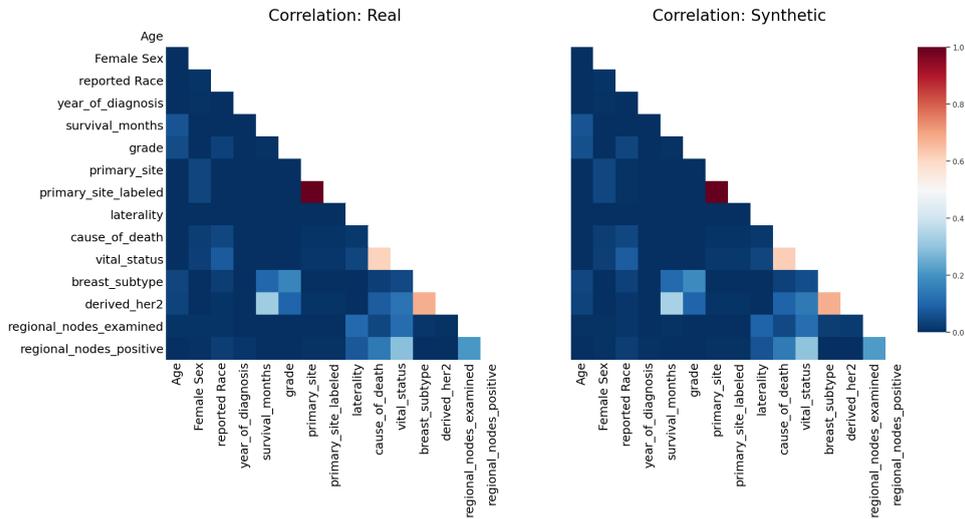

**Figure 10d:** real vs synthetic - BREAST - pairwise correlations; PCD-L1 = 0.0044

Note how the heatmaps in this case are quite similar to each other, reflecting the fact that the synthetic data maintains the pairwise correlations that exist in the original DIG dataset.

The next set of statistical fidelity metrics examine the multivariate distribution match between real and synthetic data. Concretely, in our experiments we performed this analysis for each dataset using the following: UMAP visualization, predictive performance of ML models, discriminator AUC and pMSE.

We computed the UMAP dimensionality reduction as a visual aid, comparing real to synthetic data, as shown in figures 11a-11d:

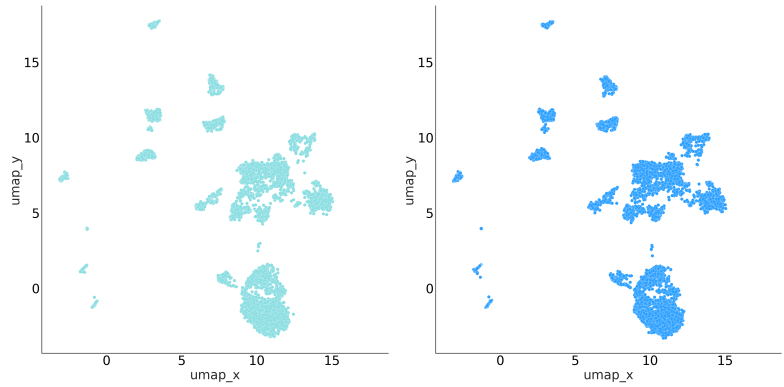

**Figure 11a:** DIG - UMAP dimensionality reduction - real vs synthetic

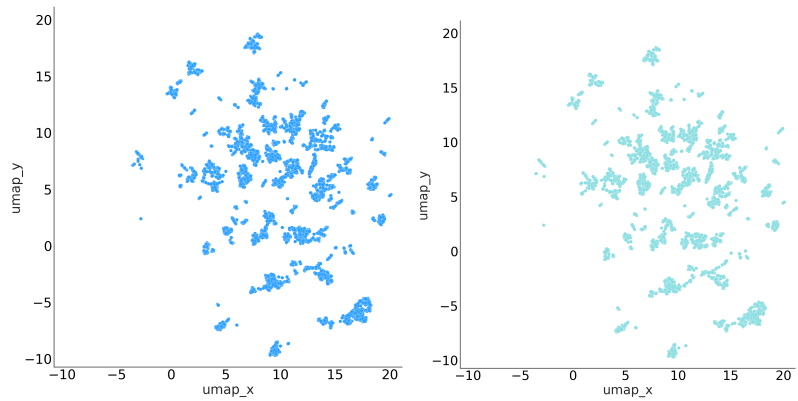

**Figure 11b:** NIS - UMAP dimensionality reduction - real vs synthetic

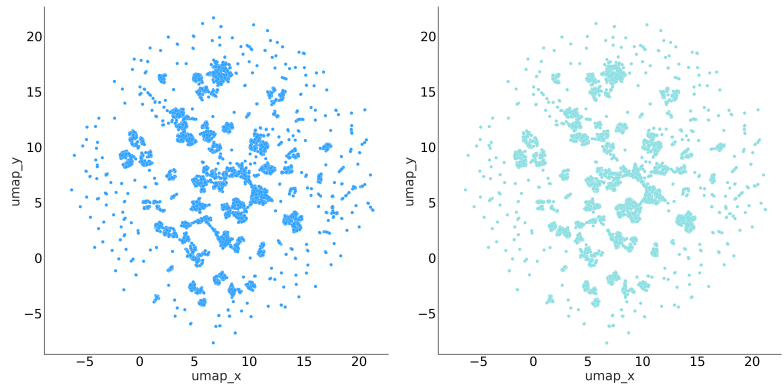

**Figure 11c:** TEXAS - UMAP dimensionality reduction - real vs synthetic

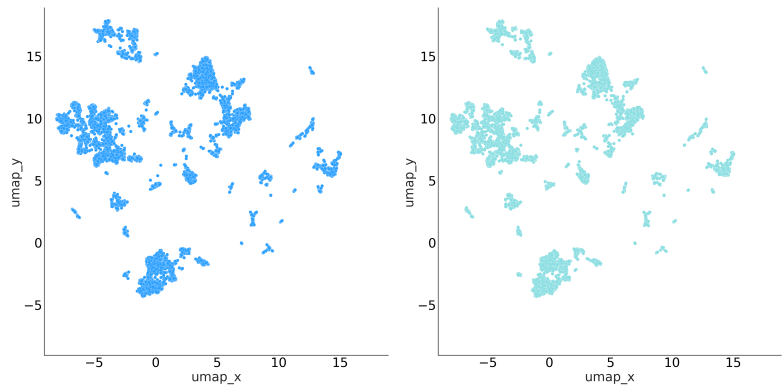

**Figure 11d:** BREAST - UMAP dimensionality reduction - real vs synthetic

This UMAP visualization demonstrates the match in coverage between real and synthetic, across all clusters of patient records, even for those who represent rare cohorts.

To further evaluate nonlinear multivariate statistical fidelity, for each dataset we trained a machine learning model (using gradient boosted trees) to predict an outcome of choice from the dataset, using selected predictors from the data. We trained this model on 80% of the real data and left 20% of the data for validation. We then trained a model on the synthetic data and validated its performance against the same validation set (20% of the real data). We repeated this 5 times, each time generating a different synthetic dataset (using random seeds) and measured the resulting metrics for each, reporting the mean and standard deviation. We also computed the feature importance of each model (using SHAP values) and compared the top-ranking features for the model trained on real data vs the model trained on synthetic data. This is demonstrated in figures 12a-12j:

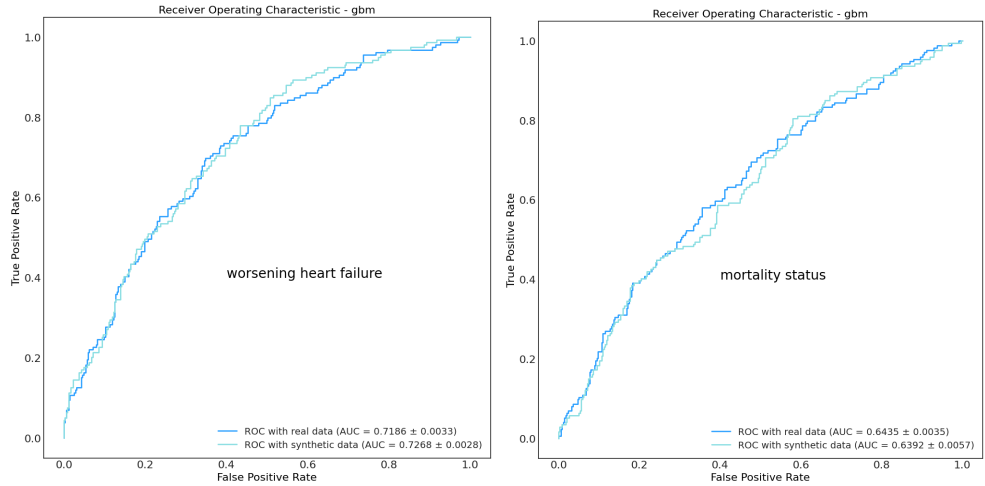

**Figures 12a and 12b:** DIG - predicting "hosp. for worsening heart failure" (left) and "all-cause mortality" (right)

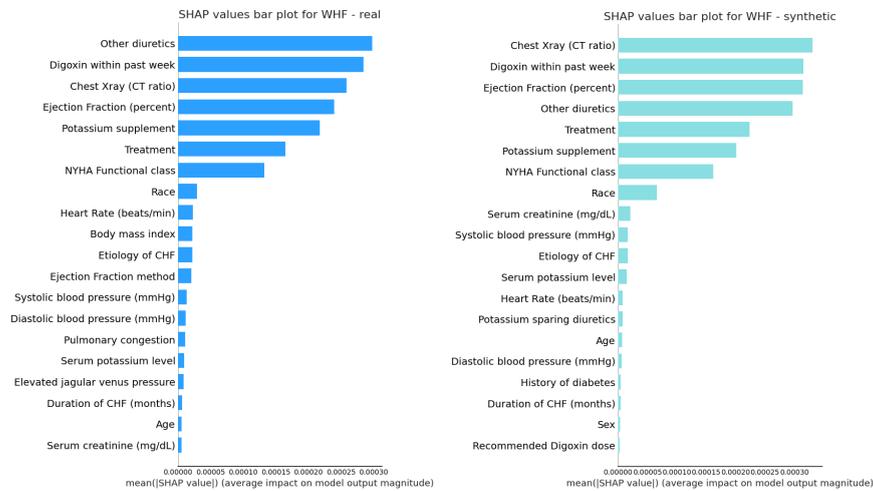

**Figure 12c:** DIG - feature importance (SHAP) for "hosp. for worsening heart failure." nDCG = 0.973

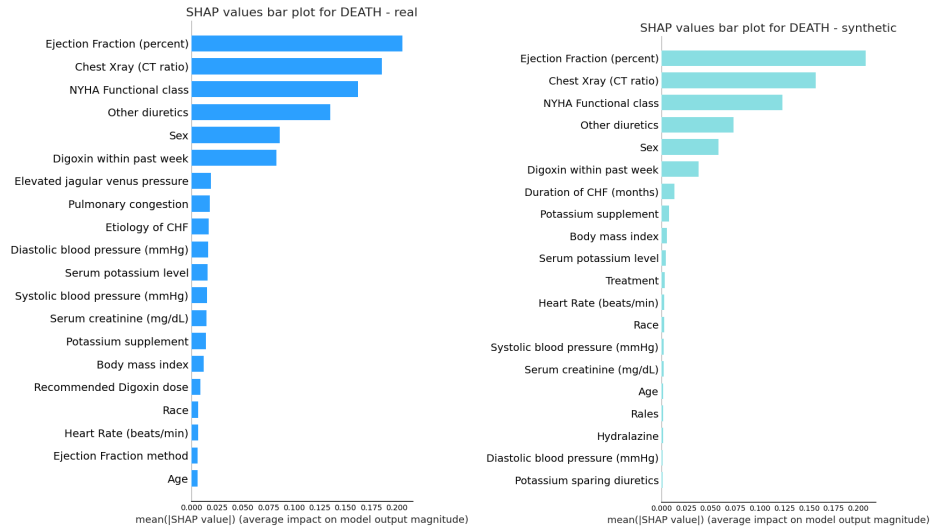

**Figure 12d:** DIG - feature importance (SHAP) for "all-cause mortality" - real vs synthetic. nDCG = 0.947

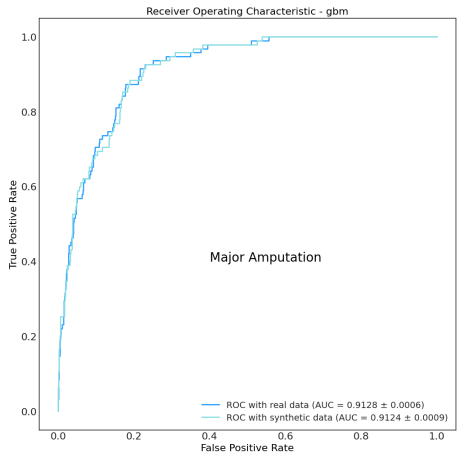
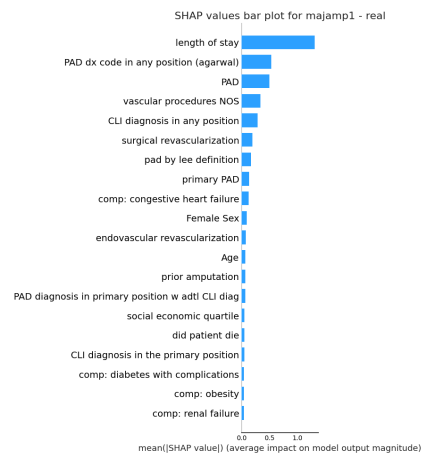
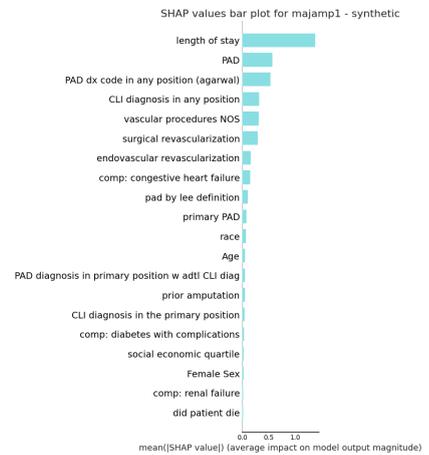

**Figures 12e and 12f:** NIS - predicting "major amputation" and feature importance (nDCG = 0.9868)

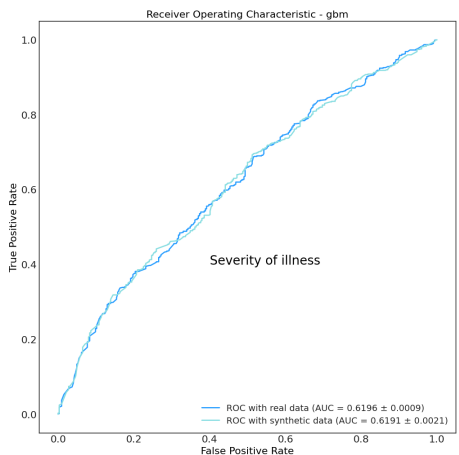
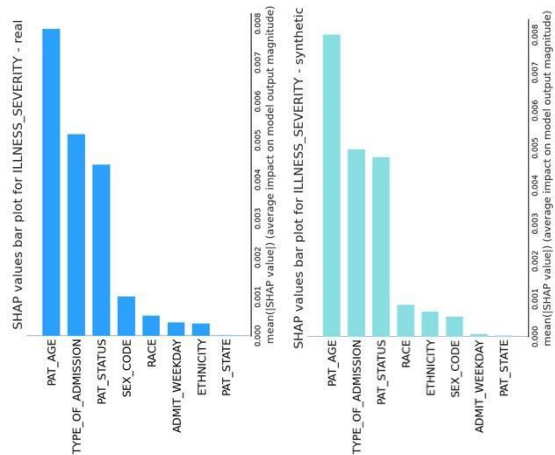

**Figures 12g and 12h:** TEXAS - "illness severity" and feature importance (nDCG=0.9819) real vs synthetic

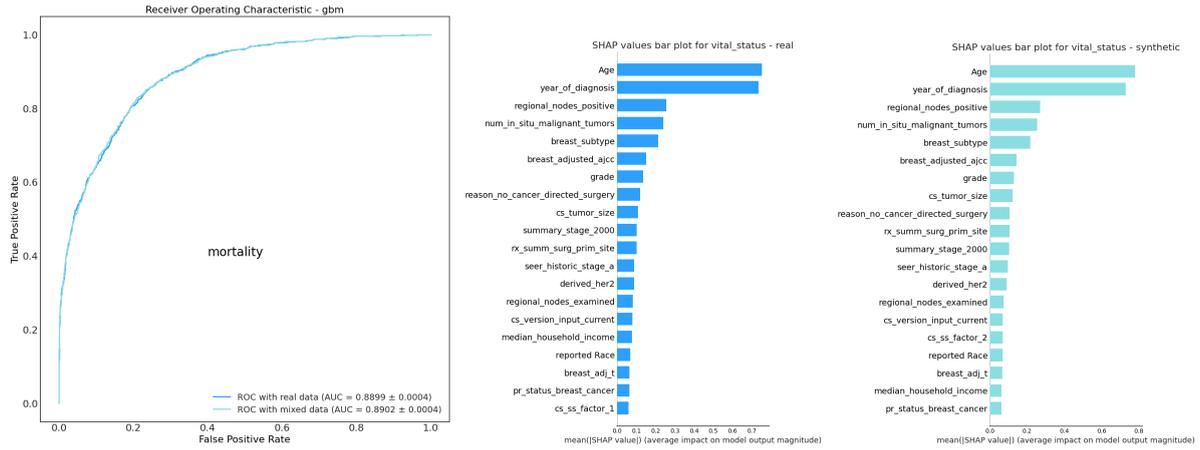

**Figures 12i and 12j:** BREAST - predicting vital status & feature importance real vs synthetic. (nDCG = 0.9969)

Two of our datasets include time-to-event variables, and for those we perform survival analysis using Kaplan-Meier as shown in figures 13a-b

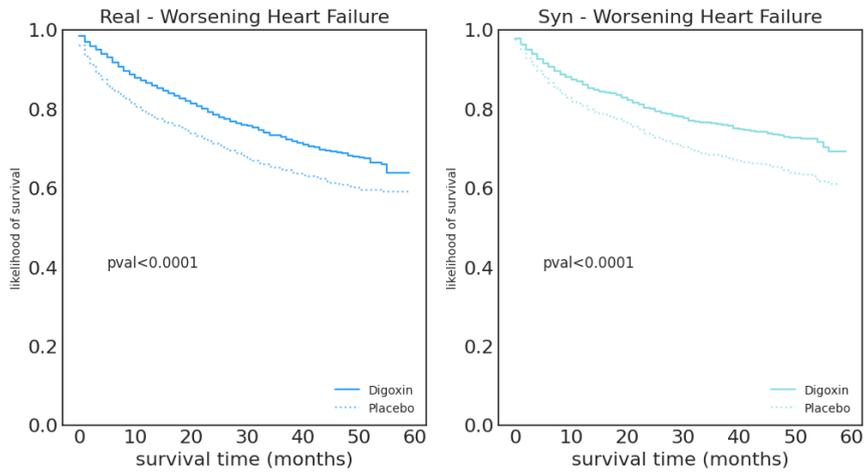

**Figure 13a:** DIG - Kaplan Meier analysis for "hospitalization for worsening heart failure" - real vs synthetic

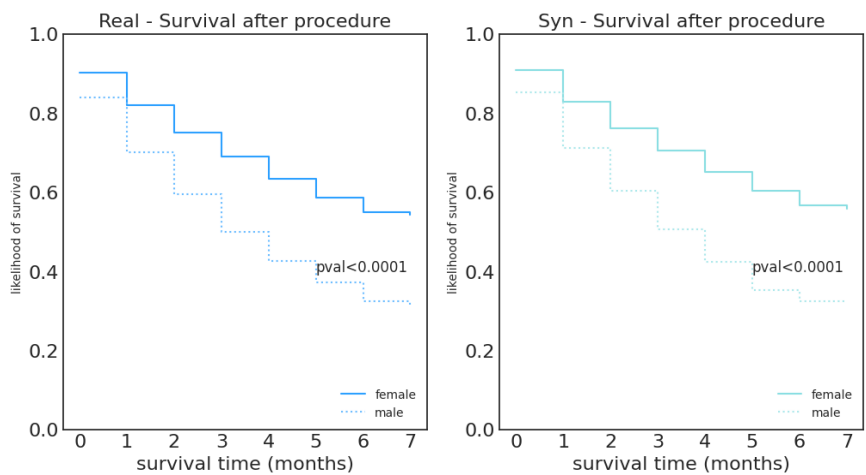

**Figure 13b:** BREAST - Kaplan Meier analysis for vital status - real vs synthetic

As described in 3.5.3, two useful metrics for a match in multivariate statistical properties between real and synthetic are the discriminator AUC and pMSE. We computed these metrics for our experimental datasets, each time comparing real to 5 different generated synthetic datasets - the results are shown in table 7:

| Dataset | DISC-AUC | pMSE |
|---------|----------|------|
| DIG | 0.6405 ± 0.0046 | 0.0706 ± 0.0004 |
| NIS | 0.5554 ± 0.0092 | 0.0193 ± 0.0002 |
| TEXAS | 0.6577 ± 0.0071 | 0.0369 ± 0.0012 |
| BREAST | 0.5406 ± 0.0025 | 0.1715 ± 0.0000 |

**Table 7:** discriminator AUC and pMSE metrics for all datasets

### 5.2.2 Privacy

As described in section 4 above, we demonstrate privacy preservation by simulating membership inference and attribute inference and use distance-to-closest-record (DCR) to demonstrate visually the distance between records in the real dataset to the closest record in the synthetic dataset, as a copy-protection measure.

In our experiments we implemented a measure of distance between records as described in 3.1. It is useful to look at the DCR distribution between real and synthetic data, as well as compare this to the distribution of DCR between real records and real records, as described above. Figures 14a-d show the resulting DCR distributions for our 4 datasets, both synthetic-to-real (green) and real-to-real (blue):

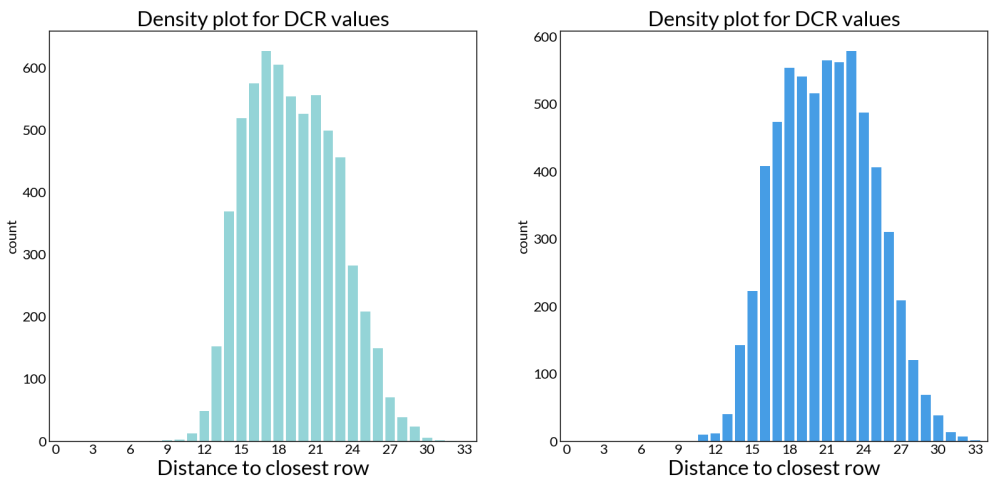

**Figure 14a:** DIG - distance to closest record synthetic-to-real (green) vs real-to-real (blue).
DCR disclosure risk = 0.0%

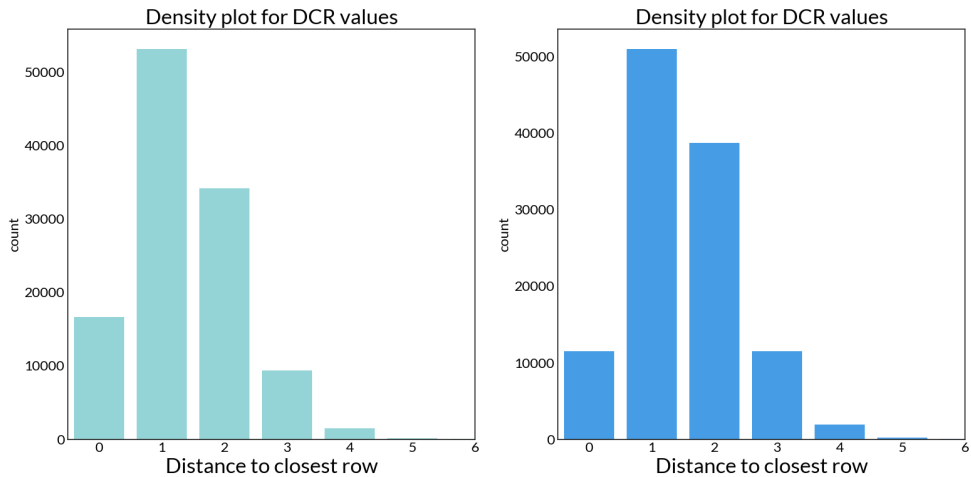

**Figure 14b:** NIS - distance to closest record synthetic vs real
DCR disclosure risk = 0.258%

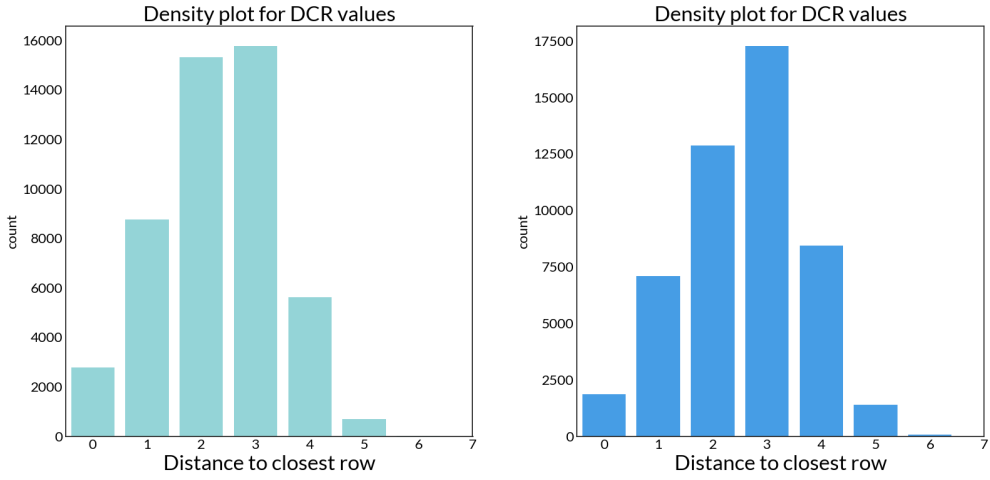

**Figure 14c:** TEXAS - distance to closest record synthetic vs real
DCR disclosure risk = 0.0%

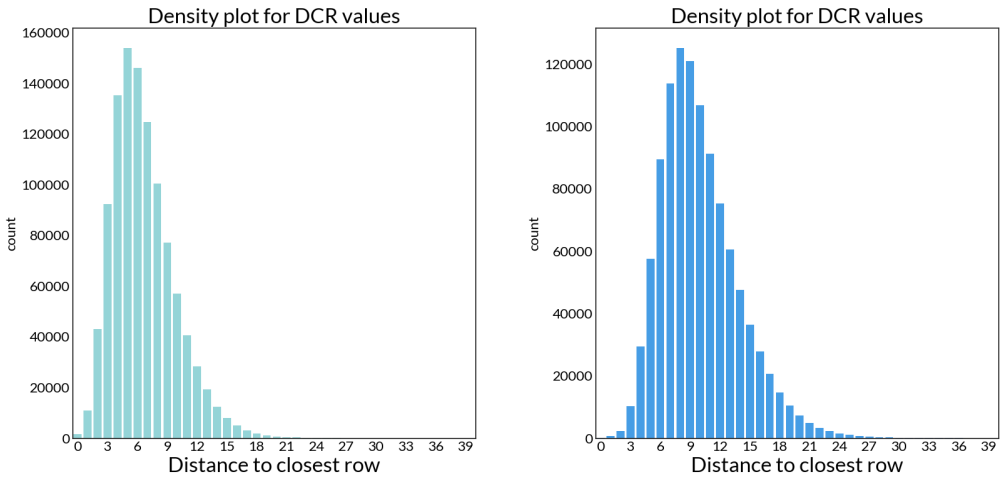

**Figure 14d:** BREAST - distance to closest record synthetic vs real
DCR disclosure risk = 0.0%

In addition to DCR, we measure in our experiments the likelihood of an adversary to successfully complete an attribute inference or membership inference attacks.

Figures 15a-d show our results for membership inference

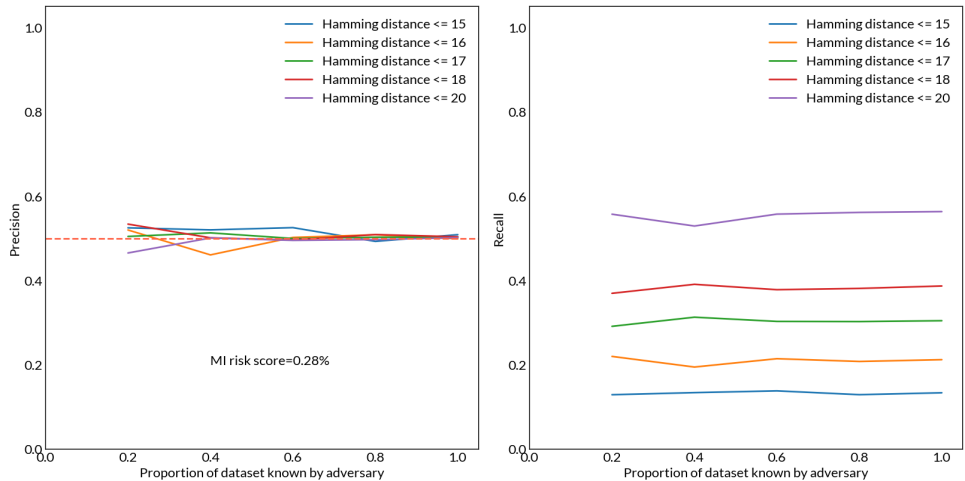

**Figure 15a:** DIG - membership inference

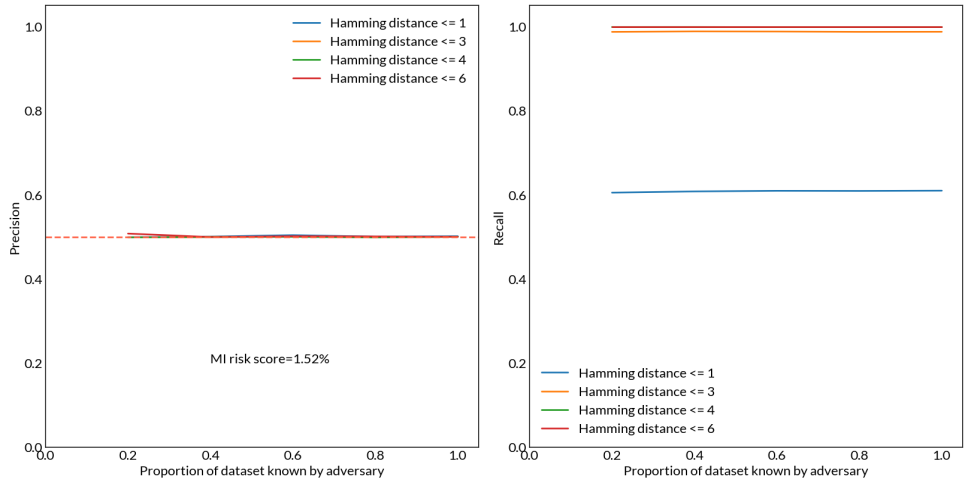

**Figure 15b:** NIS - membership inference

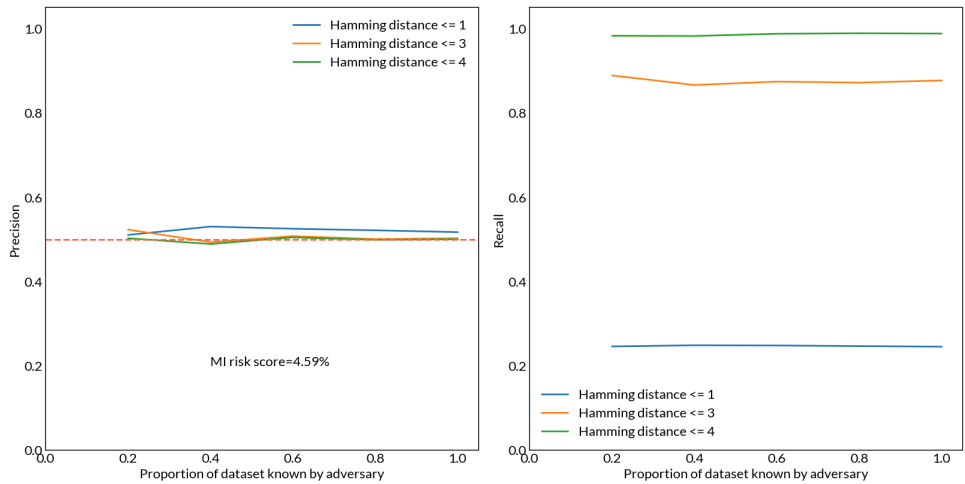

**Figure 15c:** TEXAS - membership inference

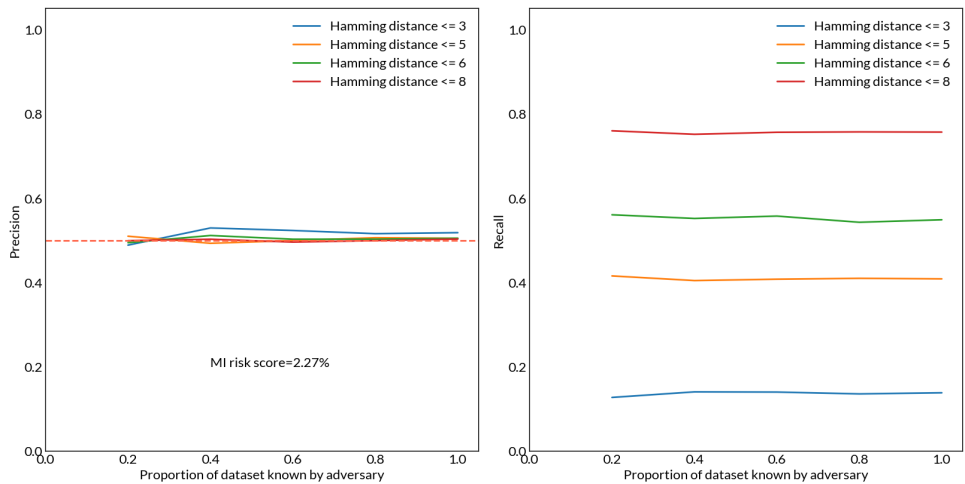

**Figure 15d:** BREAST - membership inference

Tables 8a-d show our results from the attribute inference attack simulation results:

| Disclosure risk | Death | Worsening heart failure | Number of hospitalizations | Number of symptoms | Total |
|---|---|---|---|---|---|
| 0.035588reference | 0.000195 | 0.000163 | 0.002216 | 0.0 | 0.00718 |
| paper | | | | | |
| conservative | 0.000772 | 0.000647 | 0.008776 | 0.0 | 0.028470 |
| No errors | 0.000965 | 0.000808 | 0.010970 | 0.0 | 0.035588 |

**Table 8a:** DIG - attribute inference example attack using quasi-identifiers: age, sex, and race, for the sensitive variables death, worsening-heart-failure, number of hospitalizations and number of symptoms.

| Disclosure risk | Mortality | Length of Stay | Major Amputation | Smoker | Total |
|---|---|---|---|---|---|
| reference paper | 0.0 | 0.003617 | 0.0 | 0.000375 | 0.005592 |
| conservative | 0.0 | 0.014955 | 0.0 | 0.001552 | 0.023122 |
| No errors | 0.0 | 0.018694 | 0.0 | 0.001941 | 0.028903 |

**Table 8b:** NIS - attribute inference example attack using quasi-identifiers age, gender, race and income quartile, for the sensitive variables mortality, length of stay, major imputation and smoking status.

| Disclosure risk | Illness Severity | Length of Stay | Patient Status | Risk of Mortality | Type of admission | Total |
|---|---|---|---|---|---|---|
| reference paper | 0.000257 | 0.000297 | 1.6e-5 | 9.9e-5 | 0.000320 | 0.000608 |
| conservative | 0.001020 | 0.001179 | 6.5e-5 | 0.000394 | 0.001270 | 0.002410 |
| No errors | 0.001275 | 0.001474 | 8.2e-5 | 0.000493 | 0.001588 | 0.003012 |

**Table 8c:** TEXAS - attribute inference example attack using quasi-identifiers age, race and ethnicity for the sensitive variables illness severity, length of stay, patient status, risk of mortality and type of admission.

| Disclosure risk | Vital status | Cause of death | Primary site | Tumor size | Total |
|---|---|---|---|---|---|
| reference paper | 7.5e-6 | 1.81e-7 | 2.95e-5 | 3.52e-5 | 0.000101 |
| conservative | 2.97e-5 | 7.16e-7 | 0.000116 | 0.000139 | 0.000401 |
| No errors | 3.71e-5 | 8.96e-7 | 0.000146 | 0.000174 | 0.000501 |

**Table 8d:** BREAST - attribute inference example attack using quasi-identifiers: age, race and sex for the sensitive variables vital status, cause of death, tumor size and primary site.

## 6. Discussion and Analysis

We now discuss our observations from the experimental results for each dataset.

### 6.1 DIG dataset results analysis

The DIG clinical trial dataset includes 6,800 participant records, each with 71 variables. We trained Syntegra's engine on this dataset and generated 25,000 synthetic patient records, then compared the real to synthetic using our fidelity and privacy evaluation framework.

Analysis of fidelity for individual variable distributions is described in table 4a and 5, and figures 9a and 9b. For categorical variables the KL-divergence values range from 1e-5 to 0.011, and for numeric variables the KS-statistic values range from 0.0115 to 0.0729 consistent with good correlation between the real and synthetic distributions across the board. The pairwise heatmap visualization (figure 10a) demonstrates how the pairwise correlation in the real dataset is maintained in the synthetic, consistent with the low PCD-L1 value of 0.0153.

In figure 11a we show the 2D UMAP dimensionality reduction for both real and synthetic data, demonstrating broad coverage (fidelity) and diversity (privacy); concretely, we observe that real data is split into various "clusters" of participants that are similar to each other (points are close together) and the synthetic points cover the same cluster areas; nevertheless, the synthetic points are not superimposed directly on the real points, but rather are "nearby" in the cluster - this points to strong preservation of privacy.

For multivariate fidelity, we trained two machine learning models, based on gradient boosted trees; one to predict "all-cause mortality" (the primary outcome in the study) and another to predict "worsening heart failure", a secondary outcome that was identified as significant in the study. The ROC curves and the associated AUC-ROC metric (figures 12a and 12b) demonstrate the ability to train a ML model exclusively from the synthetic data and achieve comparable results to the model trained on the real data. Interestingly, non-linear predictive models such as we have built here, were not reported in the original study results, as that technique was not possible with the technology of the time. Synthetic data can allow novel re-use of earlier clinical trial data in support of precision medicine. Figures 12c and 12d show the feature importance graphs, based on SHAP values, demonstrating that not only the predictive models arrive at similar performance, they also utilize the same features for the model in close to the same order. The feature ranking comparison metric of nDCG for DIG is 0.973 for "worsening heart failure" and 0.947 for "all-cause mortality".

The DIG study includes several time-to-event variables, which allowed us to conduct survival analysis and compare real to synthetic as shown in figure 13a - the resulting Kaplan Meier survival curves demonstrate similar likelihood of being endpoint-free for those treated with digoxin vs the placebo and a similar (low) p-value consistent with the statistical significance of treatment-related hospitalization for worsening heart failure.

As discussed in 3.5.3 we also measured discriminator AUC and pMSE as additional measures of multivariate fidelity; for the DIG dataset, the discriminator AUC is at 0.5825 and pMSE is at 0.0664, both reasonably low and consistent with our expectations for a relatively small dataset. As we'll see below, larger datasets tend to result in better values for both discriminator AUC and pMSE.

To demonstrate the privacy of synthetic DIG data, we calculated DCR and created a plot of the distribution of DCR for every real record against the synthetic dataset, as shown in figure 14a. We observe that the lowest DCR value is about 12 (out of 71 variables) which demonstrates good separation between real and synthetic; furthermore, the real-to-real (blue) distribution (removing a comparison of each real record to itself which is, by definition, zero) is very similar to real-to-synthetic, with small deviation to the right of about 1 DCR. Since there are no values of DCR=0 there are no copied records and thus our disclosure risk estimate based on DCR=0 records 0%.

We simulated a membership inference attack, the results of which are shown in figure 15a. We observe that for Hamming distance threshold of 15, the success of membership inference attack has the highest variation, with precision values ranging from 0.65 to slightly less than 0.5 (worse than random). At a recall level above 50% (hamming distance threshold of 20) we get low precision equivalent to membership inference disclosure risk score of 0.28% representing very low risk of disclosure.

Table 8a shows our results of the attribute inference attack simulation, using age, sex and race as quasi-identifiers, to infer the values of four sensitive attributes. In all scenarios the overall disclosure risk as well as the disclosure risk for each individual sensitive variable is below the 5% threshold, representing very low disclosure risk.

## 6.2 NIS dataset results analysis

The NIS dataset includes 116,009 patient records, each with 31 variables. We trained Syntegra's engine on this dataset and generated 200,000 synthetic patient records, then compared the real to synthetic using our fidelity and privacy evaluation framework.

Analysis of fidelity for individual variable distributions is described in table 4b, and figures 9c and 9d. The metrics for univariate distributions are low, as expect, both for categorical variables (with KL-divergence values of 0.0001 for "depression" and 2.6e-5 for "Major amputation"), and for numeric variables (with KS-statistic values of 0.0095 for "age" and 0.0174 for "length of stay"); the pairwise heatmap visualization (figure 10b) demonstrates very good fidelity in pairwise correlations (although correlations are general low in this dataset), consistent with a PCD-L1 value of 0.0044.

In figure 11b we show the 2D UMAP dimensionality reduction for both real and synthetic datasets, demonstrating broad coverage (fidelity) and diversity (privacy); as with the DIG dataset, we observe that real data is split into a large number of "clusters" of participants that are similar to each other and the synthetic points cover the same areas, while not being completely superimposed on the real data but rather being nearby and similar to the real points.

For multivariate fidelity, we trained a machine learning model, based on gradient boosted trees, to predict "major amputation" - one of the primary outcomes of interest in the study. The ROC curves are shown in figure 12e demonstrating AUC of 0.9128 for real and 0.9124 for synthetic - almost identical. Figure 12f shows the feature importance graphs, based on SHAP values, with good match in the ranking, and nDCG of 0.947.

Discriminator AUC is at 0.5554 and pMSE is at 0.0193, both very low demonstrating difficulty for a discriminator model to differentiate between real and synthetic records.

The DCR plot for NIS is shown in figure 14b. We observe that the lowest DCR value is zero, which could be interpreted as potentially risky. However, as discussed above, observing a very similar distribution for the real-to-real DCR plot provides the context here - in this dataset, it is quite common to find many records, in the real dataset, that have duplicate values (DCR=0), and thus it's not surprising that synthetic records also contain a similar proportion of such records. Applying our methodology, our estimated disclosure risk due to DCR=0 is 0.258%, which represents very low risk.

From our membership inference attack simulation, shown in figure 15b, we observe that high recall is achieved quickly and with a threshold of 1 recall above 0.5 is already achieved, resulting in an overall effectiveness score for membership inference of 1.52%.

Table 8b shows our results of the attribute inference attack simulation, using age, sex, race and income quartile as quasi-identifiers, to infer the values of four sensitive attributes. In all scenarios the overall disclosure risk as well as the disclosure risk for each individual sensitive variable is below the 5% threshold, representing very low disclosure risk.

## 6.3 TEXAS dataset results analysis

The TEXAS dataset includes 50,000 patient records, each with 18 variables. We trained Syntegra's engine on this dataset and generated 100,000 synthetic patient records, then compared the real to synthetic using our fidelity and privacy evaluation framework.

Analysis of fidelity for individual variable distributions is described in table 4c, and figures 19e and 9f. The metrics for univariate distributions are low, as expect, both for categorical variables (with KL-divergence values of 0.0003 for "severity of illness" and 0.0008 for "race"), and for numeric variables (with KS-statistic values of 0.0160 for "total charges" and 0.0209 for "length of stay"); the pairwise heatmap visualization (figure 10c) demonstrates very good fidelity in pairwise correlations (although correlations are general low in this dataset), consistent with a PCD-L1 value of 0.0044.

In figure 11c we show the 2D UMAP dimensionality reduction for both real and synthetic datasets, demonstrating broad coverage (fidelity) and diversity (privacy); even more than the NIS or DIG datasets, we observe here that many small clusters and good coverage by synthetic data points.

For multivariate fidelity, we trained a machine learning model, based on gradient boosted trees, to predict "severity of illness". The ROC curves are shown in figure 12g demonstrating AUC of 0.6196 for real and 0.6192 for synthetic - almost identical. We note that in this case the predictive model is not performing very well (at ROC-AUC of 0.62) but it performs similarly for real and synthetic data, which is the main exploration of our study. Figure 12h shows the feature importance graphs, based on SHAP values, with good match in the ranking, and nDCG of 1.0.

Discriminator AUC is at 0.6577 and pMSE is at 0.0369; as with DIG, we see a discriminator can have some but not great success at discriminating between real and synthetic samples.

The DCR plot for TEXAS is shown in figure 14c. Similar to the NIS dataset, we observe that the lowest DCR value is zero, but the real-to-real DCR plot shows a similar distribution, and in fact our computed disclosure risk estimate for DCR=0 records in this case is 0%.

From our membership inference attack simulation, shown in figure 15c, we observe that precision is very close to 0.5, representing low disclosure risk, and the overall membership inference effectiveness score is 4.59%.

Table 8c shows our results of the attribute inference attack simulation, using age, race and ethnicity as quasi-identifiers, to infer the values of five sensitive attributes. In all scenarios the overall disclosure risk as well as the disclosure risk for each individual sensitive variable is below the 5% threshold, representing very low disclosure risk.

## 6.4 BREAST results analysis

The BREAST dataset includes 1,072,173 patient records, each with 117 variables. We trained Syntegra's engine on this dataset and generated 1,500,000 synthetic patient records, then compared the real to synthetic using our fidelity and privacy evaluation framework.

Analysis of fidelity for individual variable distributions is described in table 4d, and figures 9g and 9h. The metrics for univariate distributions are low, as expect, both for categorical variables (with KL-divergence values of 0.0006 for "summary stage" and 0.0002 for "laterality"), and for numeric variables (with KS-statistic values of 0.0082 for "survival months" and 0.0128 for "CS tumor size"); the pairwise heatmap visualization (figure 10d) demonstrates very good fidelity in pairwise correlations, consistent with a PCD-L1 value of 0.0044.

In figure 11d we show the 2D UMAP dimensionality reduction for both real and synthetic datasets, demonstrating broad coverage (fidelity) and diversity (privacy) across all point clusters, similar to what we observed with other datasets.

For multivariate fidelity, we trained a machine learning model, based on gradient boosted trees, to predict "severity of illness". The ROC curves are shown in figure 12i demonstrating AUC of 0.8899 for real and 0.8902 for synthetic - almost identical. Figure 12j shows the feature importance graphs, based on SHAP values, with good match in the ranking, and nDCG of 0.9969.

The BREAST dataset includes several time-to-event variables, which allowed us to conduct survival analysis and compare real to synthetic as shown in figure 13b - the resulting Kaplan Meier estimator curves demonstrate similar survival likelihoods for males/females between real and synthetic.

Discriminator AUC is at 0.5406 and pMSE is at 0.1715; as with NIS, we observe that large datasets tend to result in good discriminator AUC values (close to 0.5) reflecting difficulty of the discriminator to distinguish between real and synthetic data.

The DCR plot for BREAST is shown in figure 14d. We observe that the lowest DCR value is zero, but for a very small number of records. Our computed risk of disclosure for records with DCR=0 here is 0%.

From our membership inference attack simulation, shown in figure 15d, we observe that precision is very close to 0.5, representing low disclosure risk, and the overall membership inference effectiveness score is 2.27%.

Table 8d shows our results of the attribute inference attack simulation, using age, race and sex as quasi-identifiers, to infer the values of four sensitive attributes. In all scenarios the overall disclosure risk as well as the disclosure risk for each individual sensitive variable is below the 0.05 threshold.

## 7. Conclusions

For medical data to drive innovation, PII must be removed to maintain patient privacy, at a minimum. Unfortunately, this is no longer adequate for widespread data sharing - beyond limited access to certified faculty investigators within academic institutions, such as between academic institutions and commercial third parties where therapies are most likely to be developed, and for the most sensitive variables. Despite best practices for limiting access, terms of use agreements, and other ways to assure compliance, it is widely recognized that de-identified data can be re-identified using techniques such as membership or inference attacks. Outside the US, where GDPR applies, even a de-identified or pseudo-anonymized dataset cannot be shared. A purely synthetic dataset, however, since it contains no actual patient data, is not subject to GDPR regulations.

Synthetic data has been proposed as a method which allows low-burden access to medical data, such that disclosure risk is reduced or eliminated. There is a range of methods that have been proposed by which to create synthetic data starting from a real dataset (as opposed to simulated or artificially constructed datasets based on rules). Regardless of the method, for widespread adoption, an acceptable level of fidelity and privacy must be obtained. In this paper, we have described a suite of metrics to assess a) the statistical fidelity of synthetic data as compared to the original data, and b) the disclosure risk or risk of re-identification. We further demonstrated our experimental results on four different datasets using Syntegra's Medical Mind synthetic data engine.

Other novel methods of testing privacy can be considered, such as making a synthetic dataset openly accessible to Kaggle or other crowd-sourced competitions, with a cash prize for anyone who can "hack" their way into exposing information about the real data that was used to generate the synthetic data.

Ultimately, our metrics, taken as a whole in relation to either a specific dataset or the overall process of synthetic data generation, need to drive a decision as to whether synthetic data may substitute for real world data. That decision could be anchored in analysis of whether the synthetic data is as good a representation of the real data, as the real data is representative of the entire population from which the real-world data is sampled. In other words, in order for insights generated by statistical analysis on the real data to be generalized to a wider group of patients, it is assumed that the real data is a representative sample of the larger universe of patients. In technical terms, we know the real data is a sample from an unknown probability distribution, and we expect the synthetic data to be sampled as near as possible from that same distribution.

Similarly, we believe release of high fidelity synthetic data can dramatically reduce the risk of disclosure, especially relative to traditional de-identification, but is not required to be 100% perfect privacy in order to be practically usable. To date, we have been unable to re-identify any real data based on the synthetic version, and the privacy metrics applied to synthetic data produced by the Syntegra Medical Mind suggest a very low risk of disclosure.

Once synthetic data is considered an acceptable surrogate for real data, and assuming basic access controls, a far larger group of researchers can have much quicker and less expensive access to all varieties of healthcare data. Wider access will accelerate a diverse set of insights and therapies than our current burdensome, and largely failed, system of data exchange.

### Acknowledgments

We want to thank Marlene Grenon, Greg Zahner, Patrick Baier, John Cook and Arthur Copstein for their input and support in for their insights and assistance in constructing this manuscript, and the whole Syntegra team for their dedicated work implementing the Syntegra Medical Mind and all these validation metrics.